\documentclass[journal]{IEEEtran}
\usepackage{multirow} 
\usepackage{booktabs}
\usepackage{makecell}
\usepackage{cite}
\usepackage{amsmath} 
\usepackage[ruled]{algorithm2e} 
\usepackage{array}
\usepackage{graphicx}
\usepackage{float}
\usepackage{fixltx2e}
\usepackage{amssymb}
\usepackage[table,xcdraw]{xcolor}
\graphicspath{{./figures/}}
\usepackage{url}  

\ifCLASSOPTIONcompsoc
  \usepackage[caption=false,font=normalsize,labelfont=sf,textfont=sf]{subfig}
\else
  \usepackage[caption=false,font=footnotesize]{subfig}
\fi

\begin{document}

\title{Automatic Velocity Picking Using a Multi-Information Fusion Deep Semantic Segmentation Network}

\author{Hongtao~Wang, Jiangshe~Zhang*, Zixiang~Zhao, Chunxia~Zhang,  Long~li, Zhiyu~Yang, Weifeng~Geng
\thanks{Corresponding author: Jiangshe Zhang. E-mail: jszhang@mail.xjtu.edu.cn.}
\thanks{H.T. Wang, J.S. Zhang, C.X. Zhang, Z.X. Zhao, L. Long are with the School of Mathematics and Statistics, Xi’an Jiaotong University, Xi’an, Shaanxi, 710049, P.R.China.}
\thanks{Z.Y. Yang, W.F. Geng are with the Geophysical Technology Research Center of Bureau of Geophysical Prospecting, Zhuozhou, Hebei, 072751, P.R.China.}
\thanks{The research is supported by the National Key Research and Development Program of China under grant 2020AAA0105601, the National Natural Science Foundation of China under grant 61976174 and 61877049.}}

\markboth{Journal of \LaTeX\ Class Files,~Vol.~14, No.~8, August~2015}%
{Shell \MakeLowercase{\textit{et al.}}: Bare Demo of IEEEtran.cls for IEEE Journals}

\maketitle

\begin{abstract}
Velocity picking, a critical step in seismic data processing, has been studied for decades. Although manual picking can produce accurate normal moveout (NMO) velocities from the velocity spectra of prestack gathers, it is time-consuming and becomes infeasible with the emergence of large amount of seismic data. Numerous automatic velocity picking methods have thus been developed. In recent years, deep learning (DL) methods have produced good results on the seismic data with medium and high signal-to-noise ratios (SNR). Unfortunately, it still lacks a picking method to automatically generate accurate velocities in the situations of low SNR.
In this paper, we propose a multi-information fusion network (MIFN) to estimate stacking velocity from the fusion information of velocity spectra and stack gather segments (SGS). In particular, we transform the velocity picking problem into a semantic segmentation problem based on the velocity spectrum images. Meanwhile, the information provided by SGS is used as a prior in the network to assist segmentation. The experimental results on two field datasets show that the picking results of MIFN are stable and accurate for the scenarios with medium and high SNR, and it also performs well in low SNR scenarios.
\end{abstract}
\begin{IEEEkeywords}
Stack Velocity Picking, Deep Learning, Velocity Spectrum.
\end{IEEEkeywords}

\IEEEpeerreviewmaketitle

\section{Introduction}
\IEEEPARstart{V}{elocity} analysis includes issues such as prestack time migration velocity estimation, stack velocity estimation and velocity inversion. The first two are some forms of equivalent velocity, related to the way of observation, e.g. velocity measurement datum. Prestack time migration velocity uses the root mean square (RMS) velocity, ensuring that the prestack time migration impact is ideal\cite{fomel2003velocity}. NMO velocity is the term used to describe the stack velocity, which optimizes the poststack gather. Different from them, the velocity inversion problem is reconstructing a velocity model to approximate the true subsurface properties, which is one of the most difficult velocity analysis problems\cite{li2019deep,ren2020physics,liu2021deep}. In this research, we focus on the stack velocity picking problem, which is based on prestack seismic data and employed in seismic data processing to get high-quality poststack seismic data.

There are two types of traditional approaches to pick the stack velocity based on the prestack common midpoint (CMP) gather. One method is estimating the best stacking velocity from a range of constant velocities based on the images of $t^2-x^2$ plane\cite{claerbout1978derive} or the NMO correction results of the CMP gather\cite{yilmaz2001seismic}. The other one is semblance analysis, computing velocity spectrum and finding the points on velocity spectrum with the local maximum coherence value as the velocity picking\cite{taner1969velocity}. In practice, these traditional procedures mentioned above are all semi-automatic picking methods that involve a lot of human participation. Thus, geophysicists began to explore automatic picking methods. With the advancement of machine learning, a few machine learning-based techniques for solving the automatic picking problem have been developed. Schmidt et al.\cite{schmidt1992neural} proposed an image classification method based on neural networks for each element of the velocity spectrum grid in 1992, and based on the classification result, assessed whether it was a pick-up velocity. Fish et al.\cite{fish1994neural} also proposed a similar method in 1994. Unlike above two methods, Calderó-Macías et al.\cite{calderon1998automatic} used feedback neural network (FNN) to map the CMP gather in intercept time and ray parameter domain ($\tau-p$ domain) to the predefined velocity search limits, and the FNN was trained by very fast simulated annealing (VFSA). However, as the image resolution and picking complexity of the velocity spectrum become higher, the simple neural networks described above are not sufficient to fit this mapping.

In recent years, stack velocity picking problem was transformed into a  combination optimization problem or a clustering problem. Huang et al.\cite{huang2012seismic,huang2013seismic} proposed an optimization method, which used combination optimization models on the set of the local maximum, and then searched the best combination using simulated annealing and genetic algorithm. Zhang et al.\cite{zhang2016automatic} proposed an accelerated clustering algorithm based on a time-velocity image, which is generated by local event slopes. Wei et al\cite{wei2018unsupervised}. and Chen et al.\cite{chen2018automatic} clustered the points on time as well as velocity domain, and computed the centroids of each cluster as the stack velocity. Although the unsupervised method does not require labels for training, it has obvious shortcomings. First, the definition of the objective function of the optimization method requires substantial prior knowledge. Second, the clustering method is difficult to identify noises such as multiple waves. Third, the hyperparameters of models still need to be manually adjusted for both optimization methods and clustering methods in different data domain.

Recently, the rapid development of deep learning (DL) has accelerated the application in stack velocity picking. Ma et al.\cite{ma2018automatic} proposed a regression model based on convolutional neural networks (CNN), which mapped the NMO-corrected seismic gather to the NMO velocity deviation, and then got the NMO velocity (stack velocity). Differently, Biswas et al.\cite{biswas2019estimating} used recurrent neural network (RNN) to recognize NMO velocity from CMP gather fragment images directly. Since both algorithms recognize NMO velocities in CMP gathers, the SNR of CMP gathers is required, which implies the picking results are no longer optimal for unclear trajectories.
Zhang et al.\cite{zhang2019automatic} employed You Only Look Once (YOLO), which is a state-of-the-art algorithm for object detection, to detect the energy blobs on the velocity spectrum. Additionally, long short-term memory (LSTM) network, a neural network for processing sequence data, was used for optimizing the final combination of the YOLO outputs. Unfortunately, the pickup rate is usually unstable since target information on the velocity spectrum with low SNR is difficult to be detected. 
Park et al.\cite{park2020automatic} proposed a image classification method to ensure the pickup rate, which regresses each segment split by the time domain to a velocity encode. However, since the velocity vector is a 40-bit one-hot encoding, it severely limits the upper limit of the picking accuracy. Instead of directly inputting the velocity spectrum, Fabien-Ouellet et al.\cite{fabien2020seismic} used CNN to encode the CMP gather to a mid-level feature like velocity spectrum, and then used recursive CNN and two LSTM networks to return to the principal reflection time, the RMS velocity, and the interval velocity, respectively. Although the regression results of this method were continuous values, there was no excellent result on the field data in the experimental part. Unlike the method described above, Ferreira et al.\cite{ferreira2020automatic} proposed an iterative estimation method, which is more practical and closer to the operation process of the industry. In each iteration, the prestack CMP gathers were conducted NMO correction by a set of NMO velocities, particularly which were estimated on the velocity spectrum for initialization. Then, the NMO CMP gathers were detected using a deep CNN to recognize NMO velocity deviations, which updated the NMO velocities. Regrettably, due to the NMO correction in each iteration, this method has a heavy computing overhead. Moreover, Wang et al. \cite{wang2021automatic} used the semantic segmentation network U-Net\cite{unet_2015} to pick up the RMS velocity in the velocity spectrum, achieving a good picking on the synthetic dataset.

In this paper, we propose an automatic velocity picking method using multi-information fusion network (MIFN) based on U-Net, which has been widely used in seismic data processing, e.g. velocity auto-picking\cite{wang2021automatic}, seismic data reconstruction\cite{park2021method}, seismic data interpolation\cite{fang2021seismic}, first break\cite{hu2019first}, etc. Different from Wang et al. \cite{wang2021automatic}, we add the stacked gather slice (SGS), which can assist the segment, to the input of the segmentation network to achieve spectrum pickup with low SNR. Because SGS contains the adjacent sample information as well as the velocity prior information, it improves not only the picking accuracy of the low SNR velocity spectrum, but also the robustness of picking results. Moreover, compared with the iterative method\cite{ferreira2020automatic}, our method just requires one prediction, where the computational cost has been reduced significantly. In our work, to address the instability of local estimation and balance the capacity to recognize the energy clusters with different sizes, we also propose a velocity spectrum enhance method to obtain the multi-observation spectra, which simulate that people observe the spectrum with different distances and focus. Then, the multi-observation information and the SGS information are fused as the input of the U-Net. We evaluate our approach on two field seismic data sets, using both qualitative and quantitative measures, and the experimental results illustrate that our proposed MIFN can achieve more robust and accurate picking results even in the spectrum with low SNR.

\section{Methodology}
\begin{figure}[!h]
	\centering
	\includegraphics[width=3.5in]{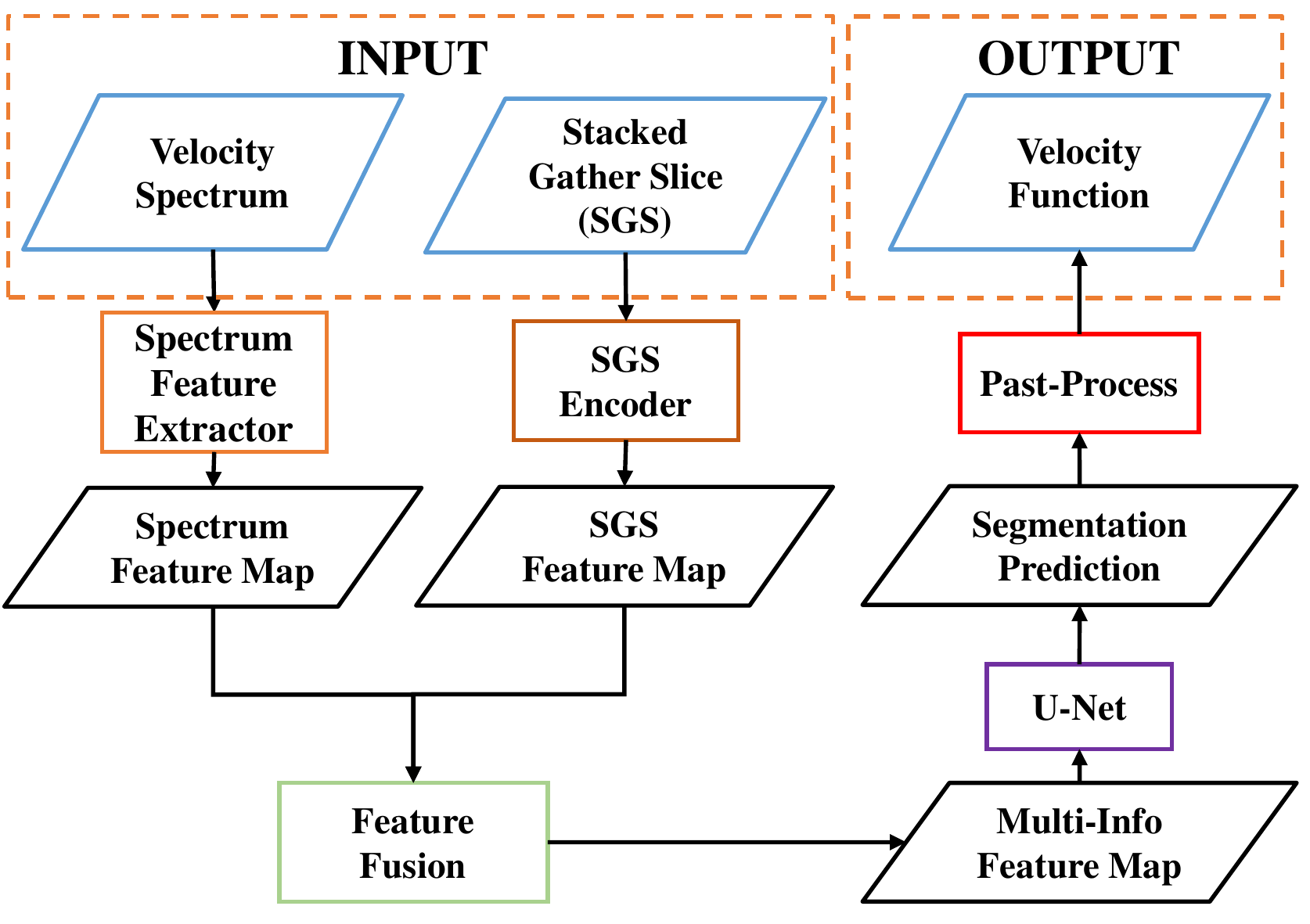}
	\caption{Workflow of the proposed.}
	\label{fig: FlowChart}
\end{figure}

The proposed velocity automatic picking method is composed of four main steps, as shown in Fig. \ref{fig: FlowChart}. First, the velocity spectrum and the SGS are input to compute feature maps separately. Second, we concatenate the feature maps from the first step to fuse the multi-information. Then, U-Net\cite{unet_2015} is employed to complete the semantic segmentation task. Finally, a post-processing for the segmentation map is exploited to obtain the final velocity function. 
\subsection{Enhanced Method of Velocity Spectrum}
Seismic velocity spectrum is calculated from CMP gathers using a coherence measure, i.e., stacked amplitude, normalized stacked amplitude, semblance, etc\cite{yilmaz2001seismic}. In this paper, we use the semblance, which is given by Eq. (\ref{semblance}), 
\begin{equation}
	N E=\frac{1}{M} \frac{\sum_{t} \sum_{i=1}^{M} f_{i, t(i)}}{\sum_{t} \sum_{i=1}^{M} f_{i, t(i)}^{2}},
	\label{semblance}
\end{equation}
where $M$ is the sample trace number in the CMP gather, and $f_{i,t(i)}$ is the amplitude value on the $i$th trace at two-way time $t(i)$, which is calculated by the stacking hyperbola using a trial velocity $v_{stk}$: 
\begin{equation}
	t(i)=\sqrt{t_{0}^{2}+\frac{x_{i}^{2}}{v_{s t k}^{2}}}.
	\label{hyperbola}
\end{equation}

To simulate the manual picking processing, we propose a spectrum enhanced approach which contains four steps:
\begin{enumerate}[\IEEEsetlabelwidth{4)}]
	\item Local smoothing: Each column of the velocity spectrum is smoothed by time window.
	\item Exponential inflation: Exponential operation for each element.
	\item Layer-wise normalization: Divide into multiple layers by line, and each layer is normalized.
	\item Amplitude limitation: Make elements less than the lower bound 0, and elements greater than the upper bound are the upper bound value.
\end{enumerate}

\begin{figure}[ht!]
	\centering
	\subfloat[]{\includegraphics[height=2in]{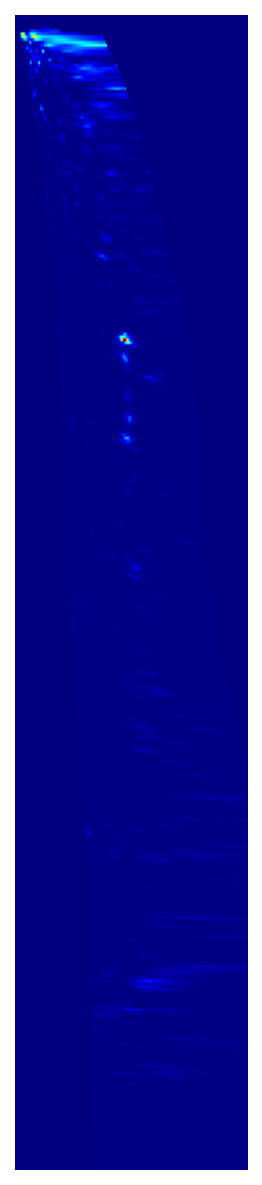}\label{fig: FE.o}}
	\hfil
	\subfloat[]{\includegraphics[height=2in]{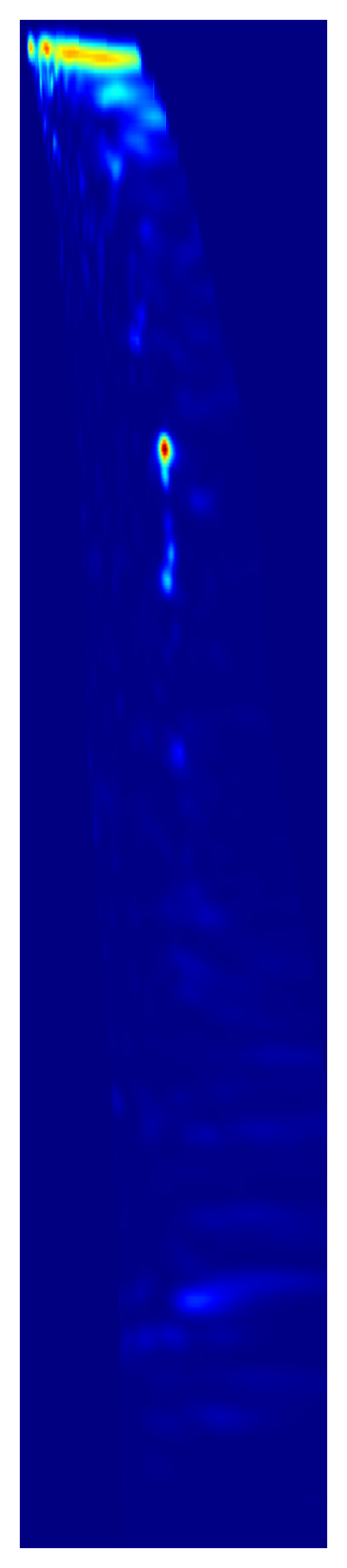}\label{fig: FE.1}}
	\hfil
	\subfloat[]{\includegraphics[height=2in]{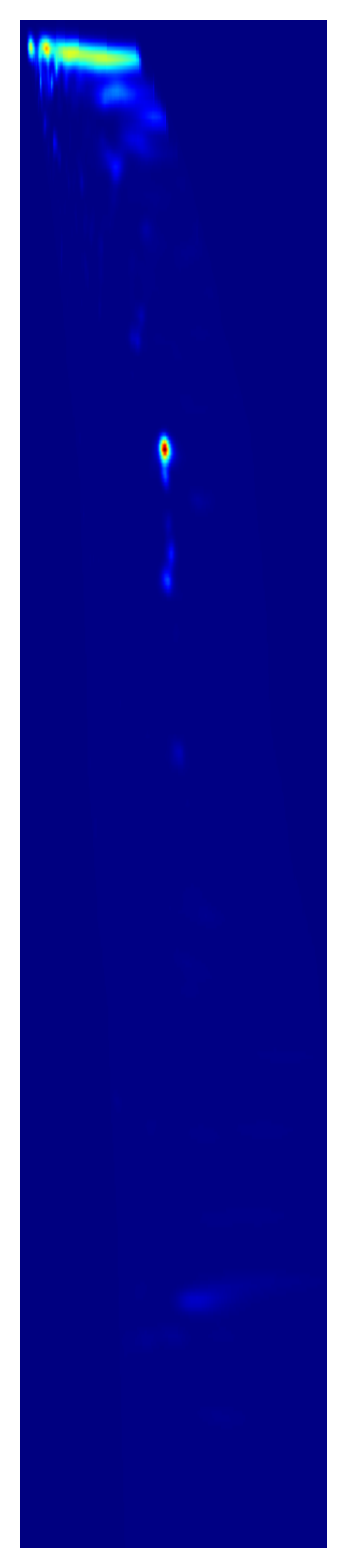}\label{fig: FE.2}}
	\hfil
	\subfloat[]{\includegraphics[height=2in]{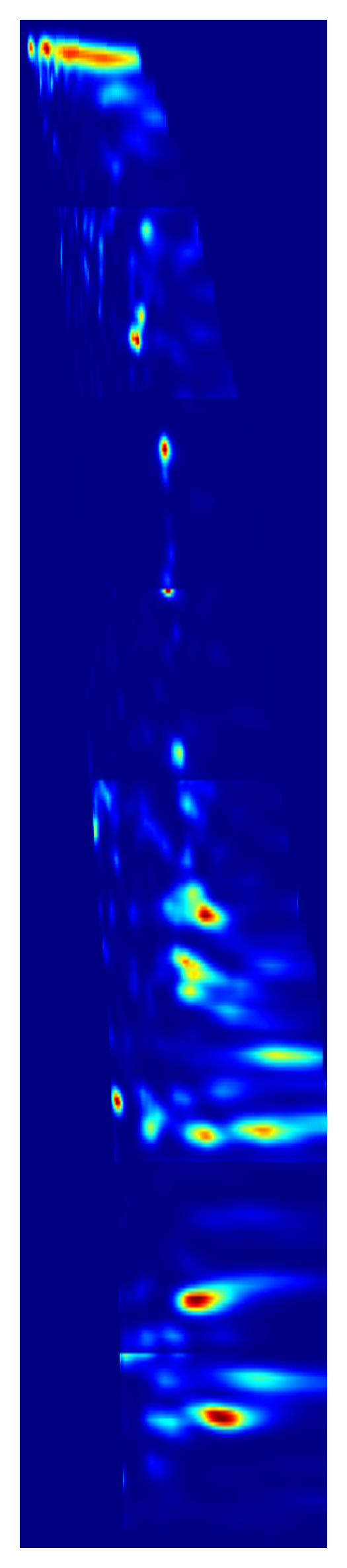}\label{fig: FE.3}}
	\hfil
	\subfloat[]{\includegraphics[height=2in]{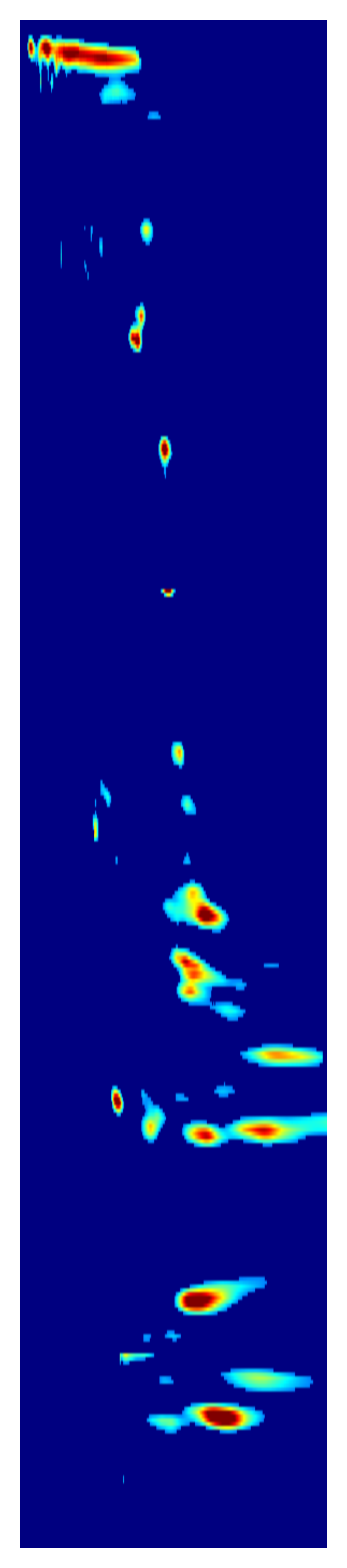}\label{fig: FE.4}
	}
	
	\caption{Spectrum Enhance Method: (a) Original velocity spectrum.  (b) The spectrum after local smoothing. (c) The spectrum after exponential inflation. (d)  The spectrum after layer-wise normalization. (e) The spectrum after amplitude limitation.}
	\label{fig: SpecEF}
\end{figure}

\begin{figure}[!h]
	\centering
	\includegraphics[width=3in]{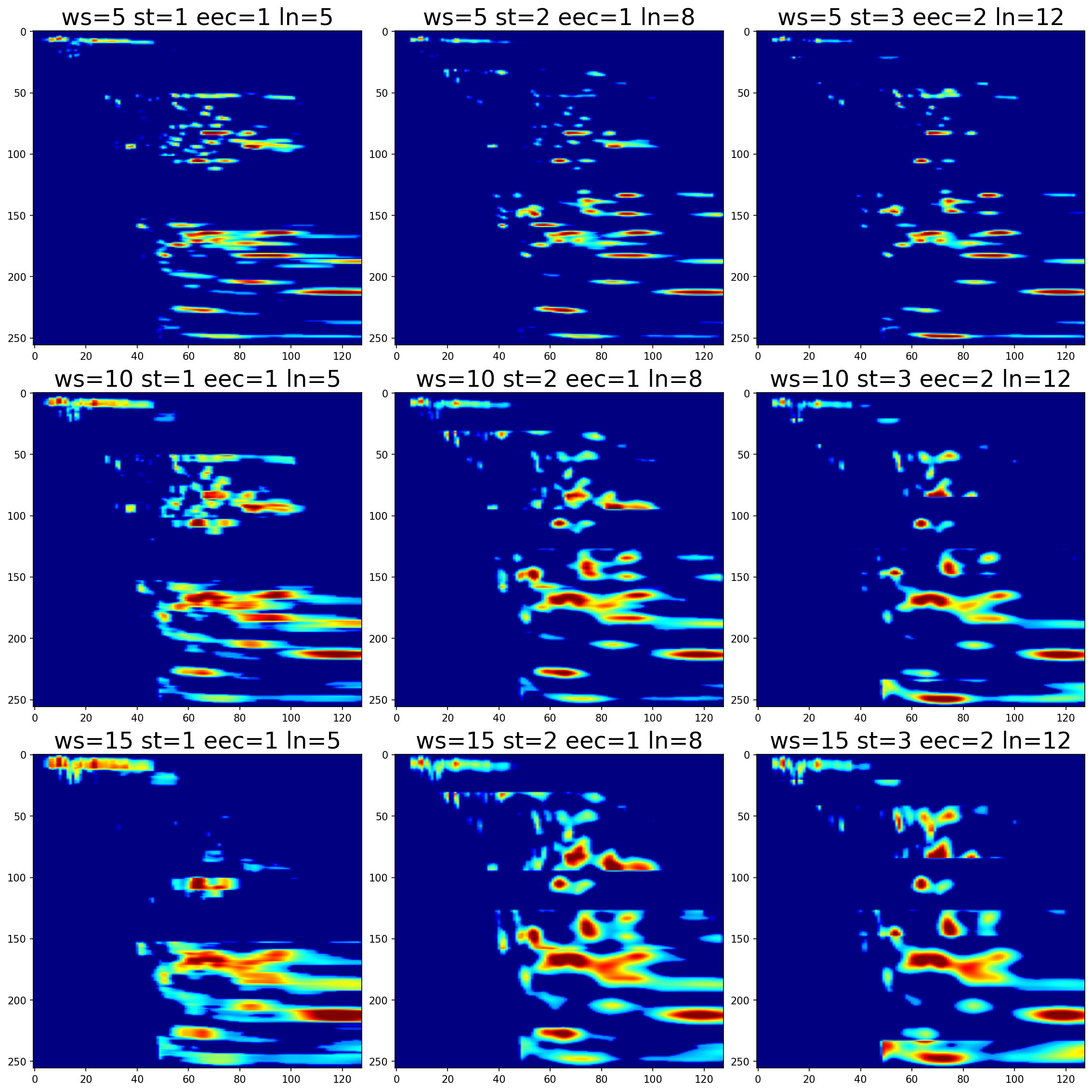}
	\caption{The Multi-scale Observation Feature Maps}
	\label{fig: FE}
\end{figure}

Fig. \ref{fig: SpecEF} shows the visualization results of the spectrum enhance method. As shown in Fig. \ref{fig: FE.1}, the local smoothing balances the column signals of the velocity spectrum, and enhances the local weak signals. Fig. \ref{fig: FE.2} implies that exponential inflation enhances the strong signal. In Fig. \ref{fig: FE.3}, layer-wise normalization balances the signals of deep and shallow layers, and greatly strengthens the weak signals in the deep layers. Finally, amplitude limitation is implemented to remove global weak signals and suppress excessively strong signals as shown in Fig. \ref{fig: FE.4}. This enhanced method equalizes the signal information of the velocity spectrum, without changing the local extremum. In addition, different combinations of the parameters involved in the enhance method can be applied to simulate the visual signals when the human eye observes the velocity spectrum at different distances with distinct focus. To balance the ability of the network to recognize energy blobs with different sizes, we select nine groups of parameters (Tab. \ref{tab: SEPara}) to generate nine enhanced feature maps (Fig. \ref{fig: FE}) as a part of network input, denoted as multi-scale feature maps, which is very helpful for velocity picking because even deep learning techniques like CNNs are difficult to learn the corresponding feature transformation. Thus, this enhanced method greatly reduces the training time cost of the DL model and improves the generalization ability of the model.

\subsection{The Encoder of Stacked Gather Slice}
For the velocity spectra with low SNR, the semantic information required for automatic velocity pickup is relatively sparse. Therefore, it is necessary to add other information to enrich the semantic information of segmentation. In this work, we choose the SGS, which contains the nearby gathers information, to assist in guiding the semantic segmentation model. In general, SGS is a stacked amplitude gather scanned by a series of stacking velocity curve, which is calculated by multiplying the reference velocity by a series of percentages, and these scanned gathers correspond to the velocity spectra of the current CMP and the adjacent CMPs. The reference velocity mentioned above generally adopts the regional velocity or the velocity of the adjacent work area. If there is no historical velocity, it is obtained by the constant velocity scan.

Specifically, the adjacent velocity spectra of the velocity spectrum $ S_{i} $ are denoted by $ S_{i1}, ..., S_{ik} $ , and the corresponding CMP gather are recorded as $ CMP_{i}, CMP_{i1}, ..., CMP_{ik} $. Using a series of stacking velocity curves $ Vc_{1}, ..., Vc_{m} $, we calculate the normal moveout (NMO) corrections for each gather, and stack the amplitude values of each row of NMO gathers. Then, the obtained $m$ SGSs are denoted by $SGS_{i1}, \cdots,SGS_{im}$ and the shape of each SGS is $ T \times k $ ($T$ is the same as the time domain of gather). $ SGS_{i1}, ..., SGS_{im} $ and  $ Vc_{1}, ..., Vc_{m} $ are all SGS information of the velocity spectrum $ S_{i} $. In particular, to ensure the consistency of the input data distribution, it is necessary to normalize the amplitudes of the gather to [-1, 1] before input the SGS encoder.

\begin{figure*}[!h]
	\centering
	\includegraphics[width=5.5in]{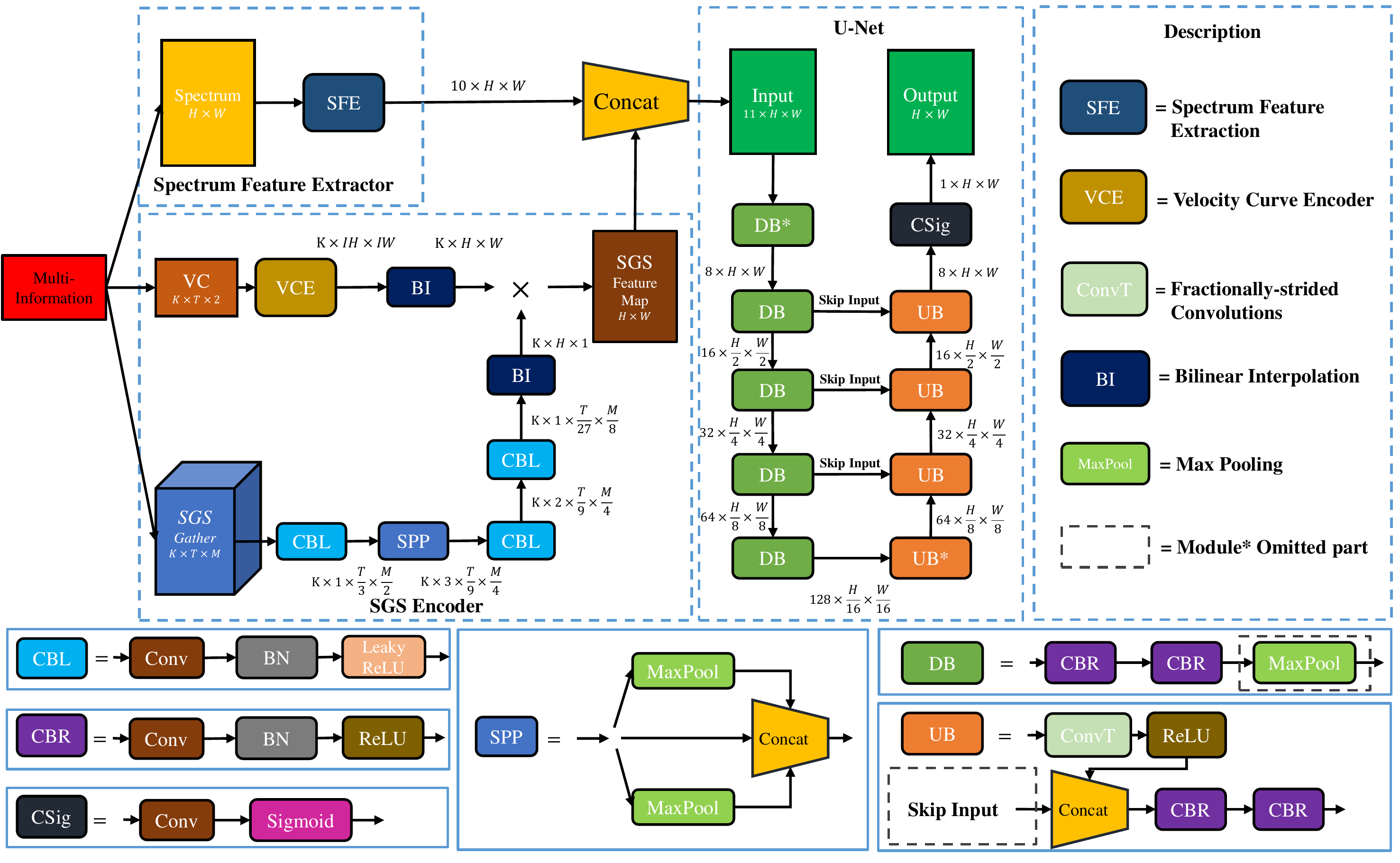}
	\caption{The flowchart of the multi-information fusion network}
	\label{fig: MIFN}
\end{figure*}

The SGS encoder consists of two encoding processing as shown in Fig. \ref{fig: MIFN}: Velocity curve is encoded as a VC mask, and SGS gather is encoded as a SGS vector. Then, each row of the VC mask is multiplied by each element of the SGS vector to obtain the SGS feature map. The VC encoder calculates $m$ masks $ M_{1}, ..., M_{m} $ from $m$ velocity curves $ Vc_{1}, ..., Vc_{m} $ respectively by the following steps:
\begin{enumerate}[\IEEEsetlabelwidth{4)}]
	\item Sampling: Sample $h$ points $ \{(t_{ijk}, v_{ijk}) | k=1,...,h \}$ at equal intervals from the velocity curve $ Vc_{j} $;
	\item Mapping: Map each point to the mask $\{Mo_{j} | j=1,...,m\}$ with the rule (eq. \ref{mapping}), where $\left[  \cdot  \right] $ is the rounding symbol, $ t_d $ and $ v_d $ are time interval and velocity interval of the axis of the mask ${Mo}_{j}$, respectively.
	\begin{equation}
		{{Mo}_{jpq}} = 
		\left\{ 
		\begin{array}{l}
				{1,\  p = \left[ {\frac{{{t_{ijk}}}}{{{t_d}}}} \right]\ and\ q =  \left[ {\frac{{{v_{ijk}}}}{{{v_d}}}} \right]}\\
				{0,\ otherwise}
		\end{array}
	    \right.
		\label{mapping}
	\end{equation}
	\item Interpolation: Using the Bilinear interpolation method, the mask $Mo_{j}$ is interpolated to a new mask $M_{j}$ of the same size as the velocity spectrum $ S_{i} $.
\end{enumerate}

The SGS gather encoder has three basic modules: convolution-batch normalization-Leaky ReLU (CBL) module, spatial pyramid pooling (SPP) module and Bilinear Interpolation (BI) module, as shown in Fig. \ref{fig: MIFN}. CBL module consists of a convolutional layer with kernel size = $5 * 3$, stride size = $3 * 2$, a batch-normalization layer, and a Leaky ReLU (Eq. \ref{LReLU}) activation layer with $\lambda=0.1$. 
\begin{equation}
	LR\left( {x,\lambda } \right) = \left\{ {\begin{array}{l}
			{x,x \ge 0}\\
			{\lambda x,x < 0}
	\end{array}} \right.
	\label{LReLU}
\end{equation}
The SPP module is proposed in 2015 to extract multi-scale feature\cite{SPP_2015}. In original SPP module, there are three max pooling layer with different kernel sizes. Due to the smaller width of SGS gathers, we only design two pooling layers with size = $ 3*3$ and $ 5*5 $. Finally, we interpolate the map to a vector with the length as same as the spectrum using Bilinear method, which has a obvious benefit of interpolation, that is flexibility to respond to dimensional changes. The size of the SGS gather may vary in practical applications, and the encoder designed by us only needs to learn the attention weight of the relative position.

\subsection{The Multi-Information Fusion Network}
In terms of network architecture, the proposed MIFN is divided into three parts: precoding part, encoder and decoder. As shown in Fig. \ref{fig: MIFN}, the precoding part extracts consists of spectrum feature extractor and SGS encoder, extracting a few basic feature maps of velocity spectrum and SGS based on geophysics knowledge. The precoded features are input to U-Net to implement semantic segmentation. After passing the encoder and decoder parts of the U-Net, we obtain the corresponding segmentation map.

The precoding part completes the information encoding and feature fusion of the velocity spectrum and SGS. To precode the information of the velocity spectrum, we use the spectrum enhance method as spectrum feature extractor (SFE) to obtain the multi-scale observation spectrum, and concatenate them together with the original velocity spectrum. There are six scale parameters in the proposed SFE: window size (ws) and smooth time (st) in step 1, exponential expansion coefficient (eec) in step 2, layer number (ln) of layer-wise normalization in step 3, and upper and lower bounds (ub \& lb) in step 4. After experiments, we select these nine sets of parameter combinations (Tab. \ref{tab: SEPara}) to generate 9 feature maps, and the ub, lb of them is 0.2, 0.8. As shown in Fig. \ref{fig: FE}, these nine feature maps simulate the focus image when people observe different layers of the velocity spectrum at different distances with distinct focus.
SGS gather and VC are encoded with SGS encoder introduced in the previous section into the SGS feature map with the same size as the velocity spectrum. Then, we concatenate the SGS feature map and the multi-scale observation spectrum to get the fusion information.
The spliced 11 multi-information fusion feature maps are the input of the segmentation network. 

\begin{table}[htbp]
	\centering
	\caption{The Parameter Combinations of Generating the Multi-scale Observation Spectrum}
	\begin{tabular}{cccccccccc}
		\toprule
		\toprule
		\textbf{No} & \textbf{0} & \textbf{1} & \textbf{2} & \textbf{3} & \textbf{4} & \textbf{5} & \textbf{6} & \textbf{7} & \textbf{8} \\
		\midrule
		\textbf{ws} & 5     & 5     & 5     & 10    & 10    & 10    & 15    & 15    & 15 \\
		\textbf{st} & 1     & 2     & 3     & 1     & 2     & 3     & 1     & 2     & 3 \\
		\textbf{eec} & 1     & 1.5   & 2     & 1     & 1.5   & 2     & 1     & 1.5   & 2 \\
		\textbf{ln} & 5     & 8     & 12    & 5     & 8     & 12    & 5     & 8     & 12 \\
		\bottomrule
		\bottomrule
	\end{tabular}%
	\label{tab: SEPara}%
\end{table}%

U-Net as a semantic segmentation network was proposed in 2015 to solve the problem of medical image segmentation. It is composed by a down-sample block (DB) and an up-sample block (UB). To keep the original semantic information and avoid vanishing gradient, skip connection is implemented between the same depth of encoder and decoder. As illustrated in Fig. \ref{fig: MIFN}, DB consists of two CBR module and a max pooling layer. CBR module is similar to CBL module, except that the last activation layer is the ReLU activation layer. The convolution kernel in CBR has the size = $3 * 3$, stride = $2 * 2$. In our work, The up-sampling part of UB is a 2D transposed convolution operator, and then input the ReLU activation layer. After concatenating the original feature maps and the up-sampled feature map, finally there are two CBR modules. At last, there are a convolution layer with kernel size = $3 * 3$, stride = $2 * 2$ and a Sigmoid activation layer. Thus, the output of the U-Net is a probability segmentation prediction map, which means that the value of each element is between 0 and 1. 

For training the MIFN, we adopt the soft label and Binary cross entropy (BCE) loss function. First, using the linear method, we interpolate the manually picked time-velocity points to obtain the velocity curve $ Vc = \left\lbrace(t_k, v_k)| k=1,...,T \right\rbrace $. Then, the velocity curve $ Vc $ is mapped to the soft label mask $ M $ of shape $=\left( T, V\right) $ with Eq. (\ref{SoftLb}):
\begin{equation}
	{M_{p,q}} = \left\{ {\begin{array}{l}
			{1,\ p = \left[ {\frac{{{t_k}}}{{{t_d}}}} \right]\ and\ q = \left[ {\frac{{{v_k}}}{{{v_d}}}} \right]}\\
			{0.8,p = \left[ {\frac{{{t_k}}}{{{t_d}}}} \right] \pm 1\ and\ q = \left[ {\frac{{{v_k}}}{{{v_d}}}} \right] \pm 1}\\
			{0,\ otherwise},
	\end{array}} \right.
	\label{SoftLb}
\end{equation}
where $\left[  \cdot  \right] $ is the rounding symbol, $ t_d $ and $ v_d $ are time interval and velocity interval of the axis of the mask $M$.
Since the prediction map of U-Net is between 0 and 1, it is not necessary to set softmax layer before calculate BCE loss function (Eq. (\ref{BCELoss})), where $ BS $ is the size of each batch, $ H $ and $ W$ are the height and the weight of the mask $M$.
\begin{equation}
	\begin{array}{l}
		{BCELoss\left( {p,t} \right) =  - \frac{1}{{BS}}\sum\limits_{n = 1}^{BS} {{l_n}\left( {{p_n},{t_n}} \right)} }\\
		{\begin{array}{*{20}{c}}
				{{l_n}\left( {{p_n},{t_n}} \right) = \frac{1}{{H \cdot W}}\sum\limits_{i = 1}^H {\sum\limits_{j = 1}^W {{t_{{n_{ij}}}} \cdot \log {p_{{n_{ij}}}}} } }\\
				{\left. { + \left( {1 - {t_{{n_{ij}}}}} \right)\log \left( {1 - {p_{{n_{ij}}}}} \right)} \right]}
		\end{array}}
	\end{array}
	\label{BCELoss}
\end{equation}
Other training details are explained in the Sec.III.B.

\subsection{Post-processing Method} 
Based on the segmentation map of MIFN, we apply the following post-processing to get the final estimated velocities. First, we find the column index of the maximum of each row on the segmentation map, and note the index pairs of column and row as $ \{(i_k, j_k) | k=1,..., V\} $, where $V$ is the row number. If there occur at least two indices corresponding to the maximum value in the same row, we take their average and convert them into integers. Then, these index pairs are mapped to the t-v points with the original scale which is corresponding to the axis of velocity spectrum. Finally, we interpolate the scaled t-v points to the velocity curve with linear method. In particular, unlike the linear extrapolation used in \cite{wei2018unsupervised,chen2018automatic}, in order to make predictions robustly for the unsegmented parts of the shallow and deep layers, we adopt a linear regression method to estimate the stack velocity of the shallow and deep layers, that is, take the points within the $T_t$ range of the shallow and deep layers for the least squares linear estimation.
 
\subsection{Quality Control}
To control the automatic picking quality of MIFN, the following two evaluation techniques are employed. Quantitatively, we focus on the average absolute error of the two velocity function at each discrete time point, and define velocity mean absolute error (VMAE) as follow:
\begin{equation}
	\text{VMAE} = \frac{1}{N}\sum\limits_{i = 1}^N {\left| {{V_{a}}\left( {{t_i}} \right) - {V_{m}}\left( {{t_i}} \right)} \right|},  
	\label{VMAE}
\end{equation}
where $N$ is the number of the discrete time, ${V_{a}}\left( {{t_i}} \right)$ and ${V_{m}}\left( {{t_i}} \right)$ are the velocities of the automatic picking and the manual picking at $t_i$.

For subjective evaluation, there are two methods to evaluate the picking results. First, we observe the quality of CMP gathers after NMO correction, which can be measured by the horizontal sum of amplitude values. Second, we compute the stacked CMP gather from all CMP gathers after NMO correction which are on the same work line, and check the texture continuity of the stacked CMP gather.

\section{Experiments}
\subsection{Datasets}
To verify the effectiveness and accuracy of our model, we selected two field prestack data sets with different SNRs, which are both from Xinjiang Tarim Oilfield in western China, and recorded as datasets A and B. As shown in Fig. \ref{fig: A-Spec} and Fig. \ref{fig: A-CMP}, in dataset A, the velocity spectrum is with medium and high SNRs, and there are a few clearer tracks in the prestack CMP gather. Fig. \ref{fig: B-Spec} and \ref{fig: B-CMP} show that in dataset B, there are no clear energy cluster on the spectrum, and the amplitude signals of CMP gather is disordered. As shown in Tab. \ref{tab: datasets}, there are 10 lines, total 4456 spectra in dataset A and 20 lines, total 12005 spectra in dataset B. Additionally, these are 1085 and 1209 velocity spectra in datasets A and B, with the labeled stack velocity called seed points. These seed points are uniformly distributed across all sampling points.

\begin{table}[ht!]
\centering
\caption{The Description of Field Datasets A and B}
\label{tab: datasets}
\resizebox{\linewidth}{!}{ 
\begin{tabular}{ccccc}
\toprule\toprule
\multirow{2}{*}{Dataset} &
  \multirow{2}{*}{\begin{tabular}[c]{@{}c@{}}Sample Number\\ (Labeled Number)\end{tabular}} &
  \multicolumn{3}{c}{Labeled Dataset Divide by Lines} \\ \cline{3-5} 
 &  & Train & Validation & Test \\ \hline
A &
  4456(1085) &
  \begin{tabular}[c]{@{}c@{}}15 lines,\\ No.2240-2800\\ (612 samples)\end{tabular} &
  \begin{tabular}[c]{@{}c@{}}5 lines,\\ No.2840-3000\\ (215 samples)\end{tabular} &
  \begin{tabular}[c]{@{}c@{}}6 lines,\\ No.3040-3240\\ (258 samples)\end{tabular} \\
B &
  12005(1209) &
  \begin{tabular}[c]{@{}c@{}}16 lines,\\ No.220-670\\ (792 samples)\end{tabular} &
  \begin{tabular}[c]{@{}c@{}}5 lines,\\ No.700-820\\ (190 samples)\end{tabular} &
  \begin{tabular}[c]{@{}c@{}}6 lines,\\ No.850-1000\\ (227 samples)\end{tabular} \\ \bottomrule\bottomrule
\end{tabular}
}
\end{table}

\begin{figure}[!h]
	\centering
	\includegraphics[width=3in]{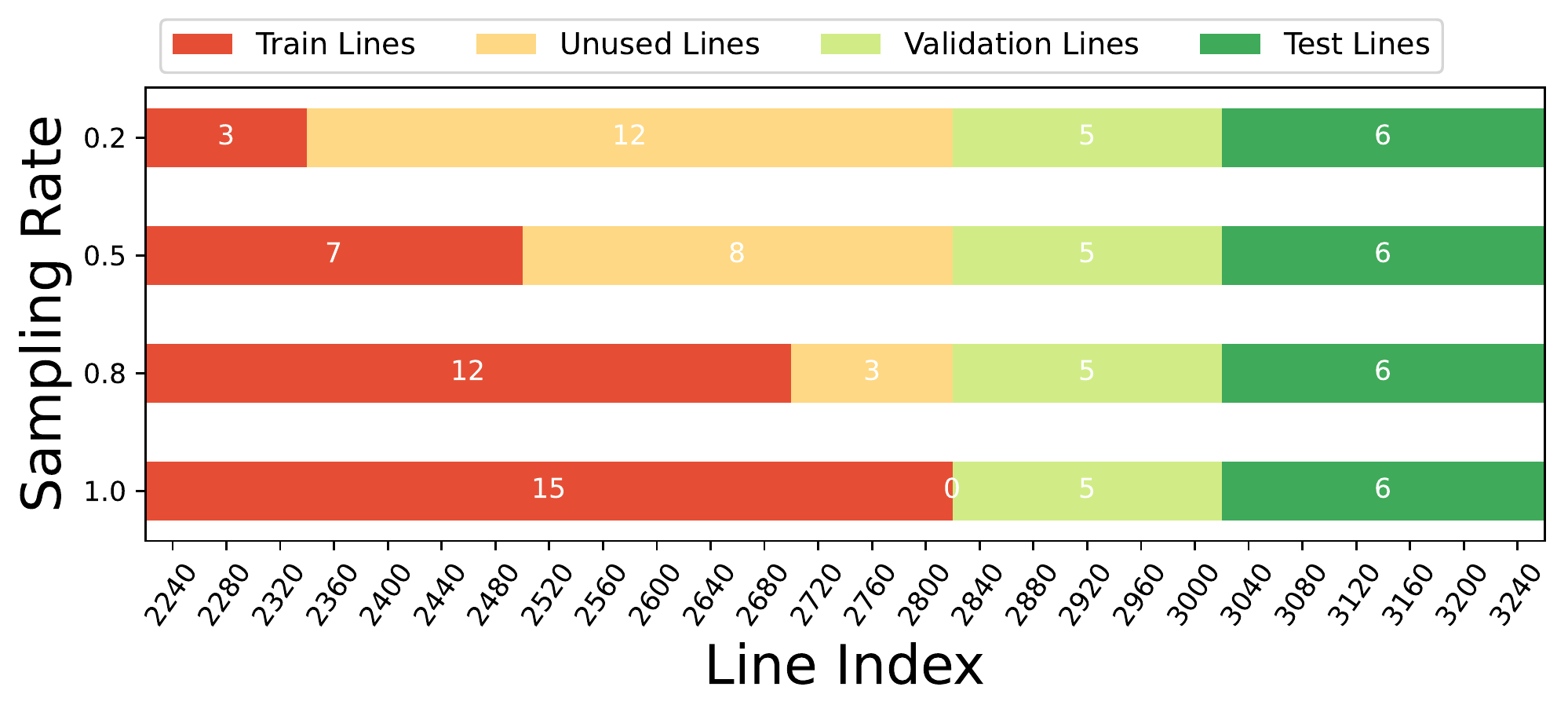}
	\caption{The split lines distribution of train set, unused set, validation set and test set in data set A.}
	\label{fig: dist-A}
\end{figure}

\begin{figure*}[ht!]
	\centering
	\subfloat[]{\includegraphics[height=2.3in]{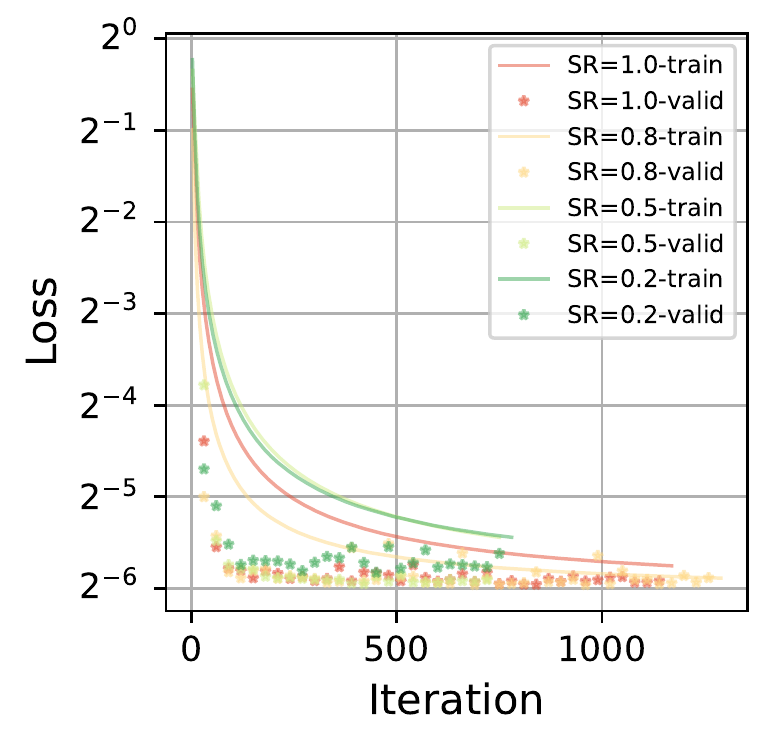}\label{fig: LA}}
	\hfil
	\subfloat[]{\includegraphics[height=2.3in]{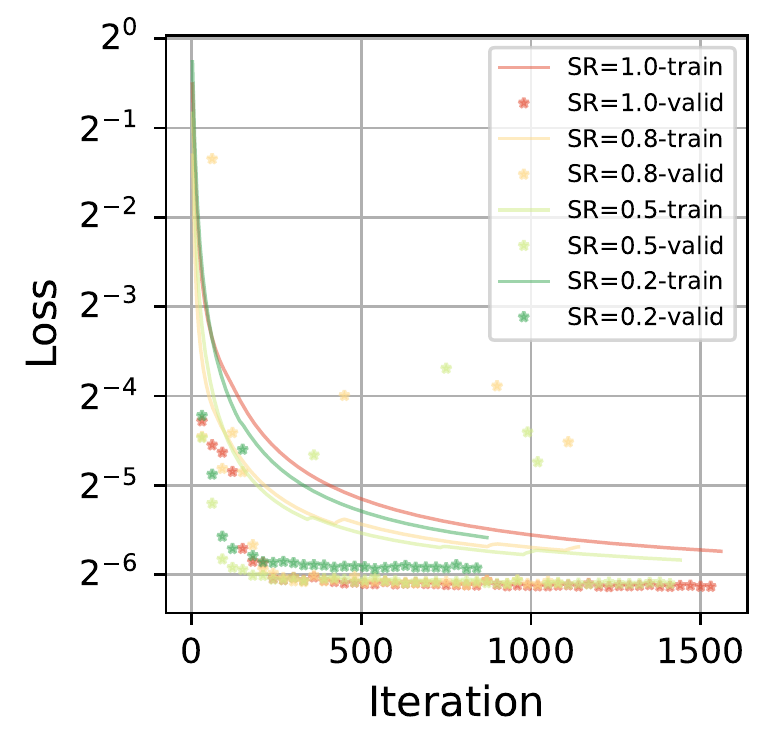}\label{fig: LB}}
	\hfil
    \subfloat[]{\includegraphics[height=2.3in]{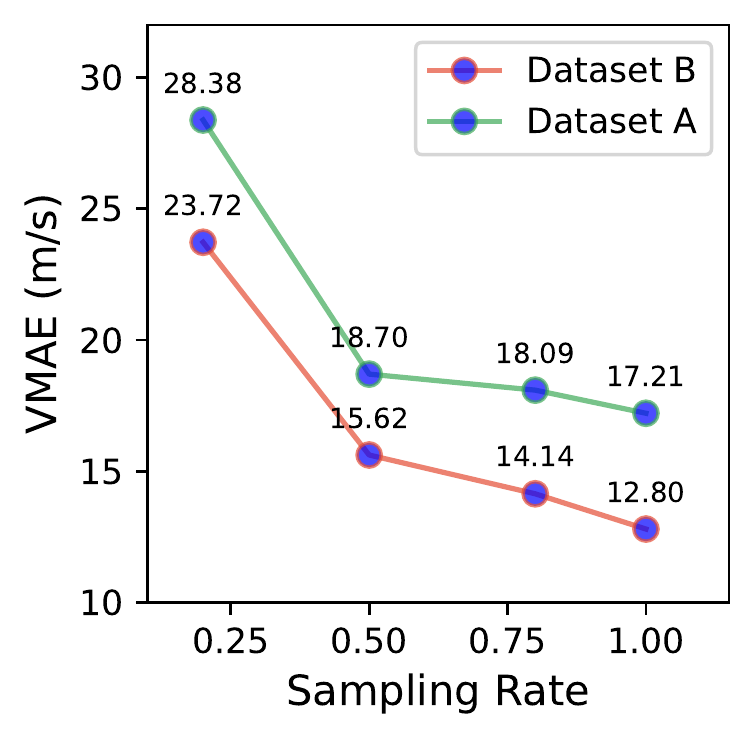}\label{fig: ABAll}}
	\hfil
	\caption{The BCE loss of validation set and train set, and VMAE of test set with different sampling rate(SR). (a) and (b) are the train and validation set BCE loss of sets A and B, respectively. In particular, the batch size of each iteration is 32. (c) is the VMAE of datasets A and B on test dataset under different sampling rate(SR).}
	\label{fig: SeedRateFig}
\end{figure*}

\subsection{Model Training}
For evaluating our MIFN on the datasets A and B, we split the seed points in each dataset into training, validation and test set with the rates of 60\%, 20\% and 20\% by sampling the lines, as shown in Tab. \ref{tab: datasets}. In particular, different from the test set, the validation set is used to verify whether the model trained by the training set are overfitting, and do not participate in the final accuracy test. Before input to MIFN, the spectrum images and the soft label masks were resized to $256 * 128$ by using Bilinear interpolation. In our work, we chose 32 as training batch size, adaptive moment estimation (Adam)\cite{Adam_2014} as the optimizer, and 0.01 as the initial learning rate. We used a computer equipped with a RTX 2080 Ti graphic card and a RTX 3090 graphic card for training. Moreover, in all the following model training process, we set 5000 as the maximum number of batch iteration and validate the model every 15 iterations, and adopt the strategy of early stop to train the model, which breaks the training iteration when the validation loss does not drop ten times consecutively.

\subsection{Results}
The results of our experiment are divided into four parts. First, the ablation experiment based on data set A aims at verifying the necessity of each main components. Second, the result of generalization ability test shows the feasibility of our model on the field data. Then, we study the performance of MIFN on the test sets of datasets A and B. Finally, fine-tuned model test is conducted to evaluate the transfer ability and practicability of MIFN.

\begin{figure*}[ht!]
	\centering
	\subfloat[]{\includegraphics[height=2.8in]{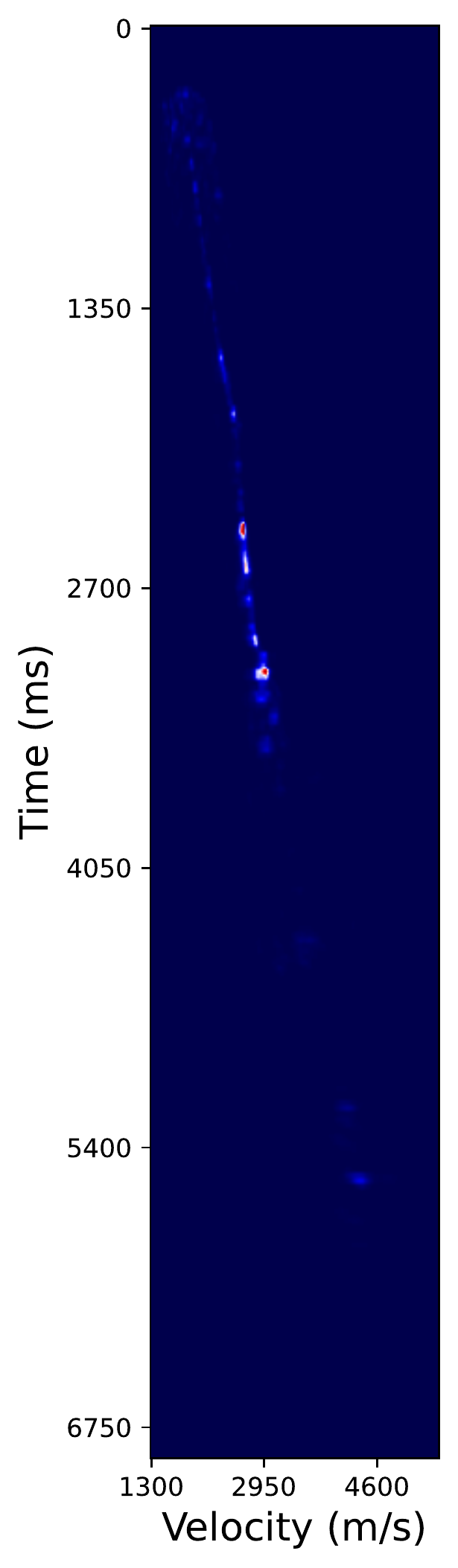}\label{fig: A-Spec}}
	\hfil
	\subfloat[]{\includegraphics[height=2.8in]{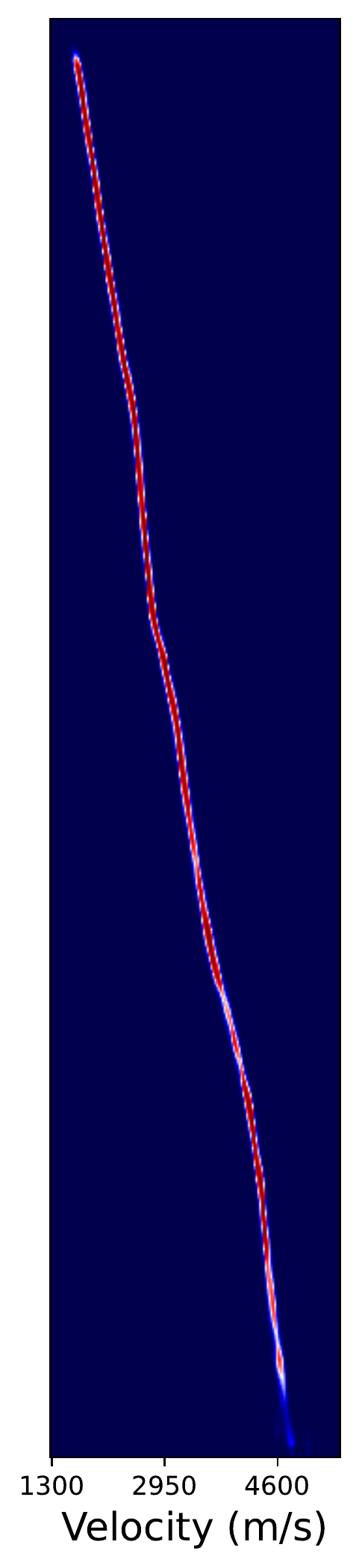}\label{fig: A-seg}}
	\hfil
	\subfloat[]{\includegraphics[height=2.8in]{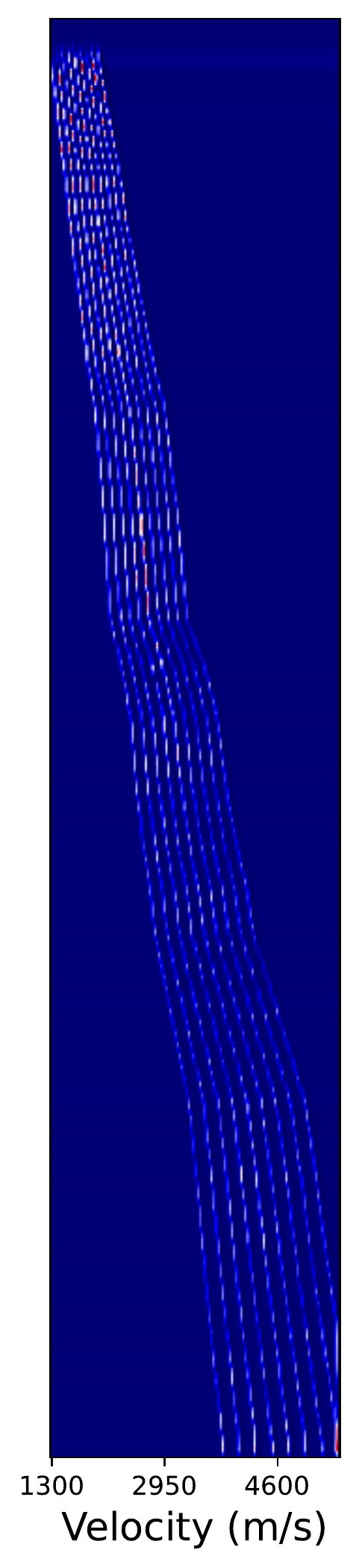}\label{fig: A-SGS}}
	\hfil
	\subfloat[]{\includegraphics[height=2.8in]{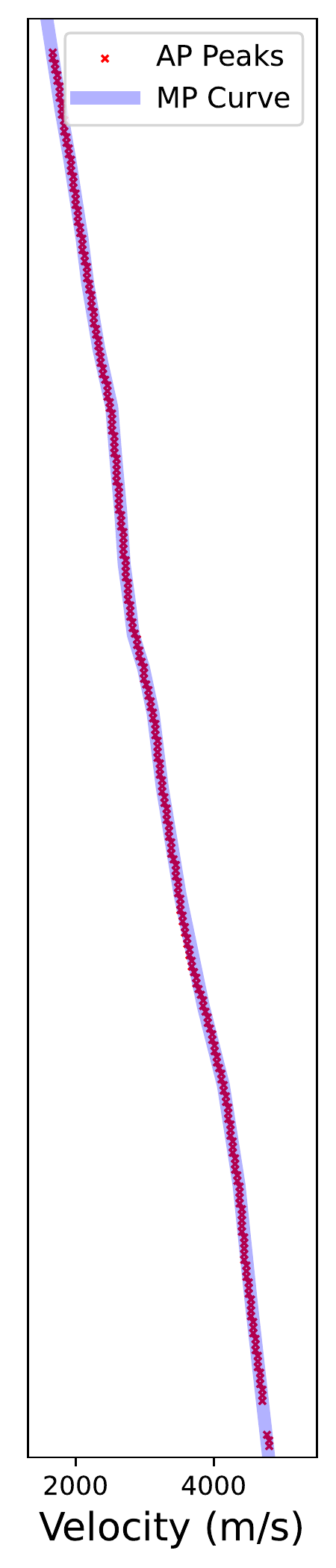}\label{fig: A-Vel}}
	\hfil
	\subfloat[]{\includegraphics[height=2.8in]{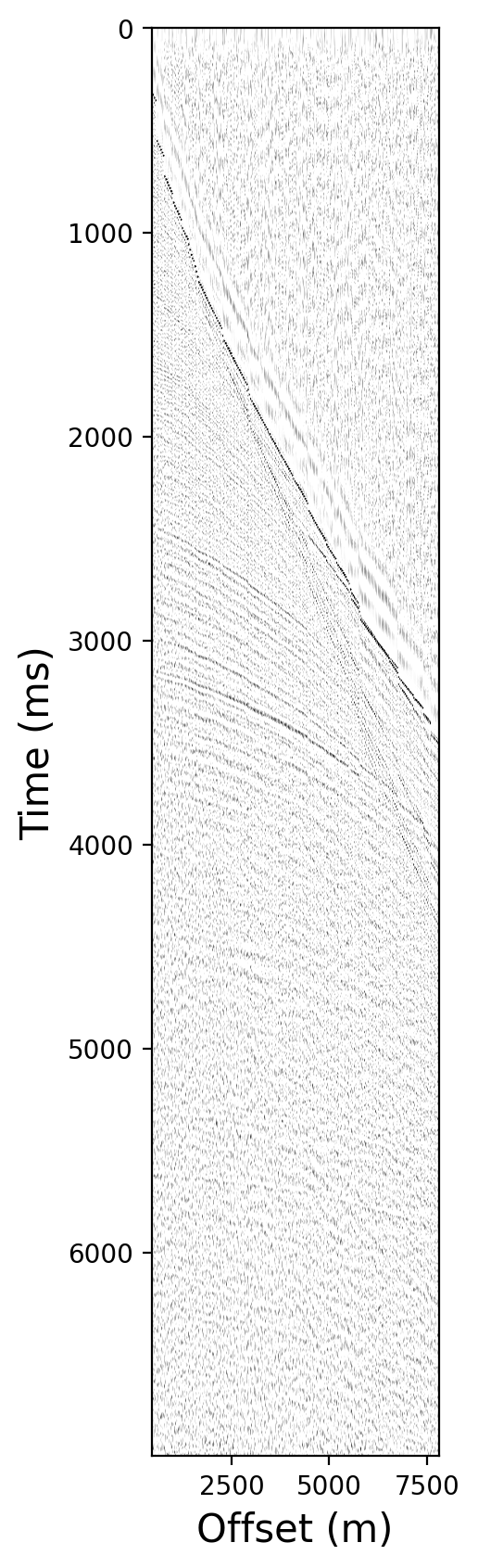}\label{fig: A-CMP}}
	\hfil
	\subfloat[]{\includegraphics[height=2.8in]{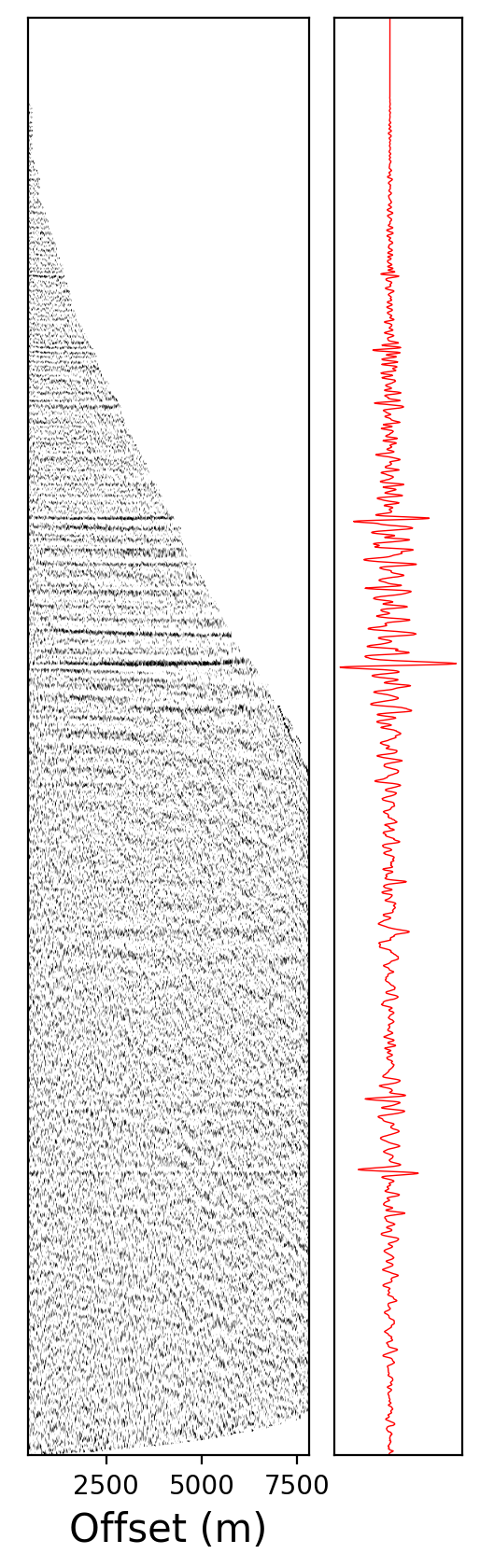}\label{fig: A-NMO}}
	\\
	\subfloat[]{\includegraphics[height=2.8in]{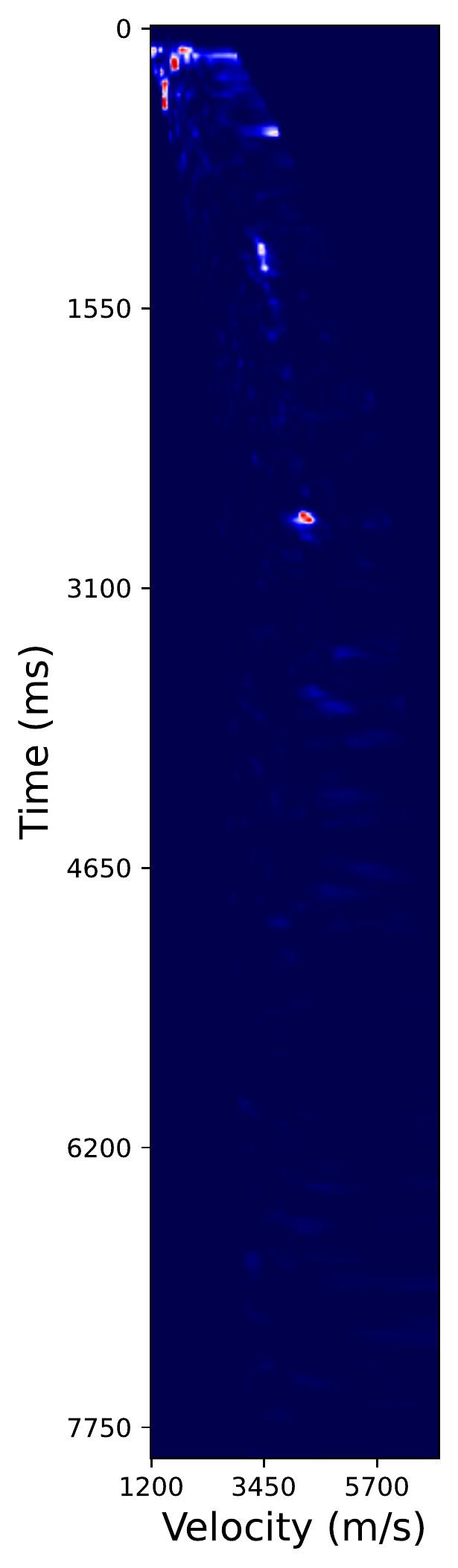}\label{fig: B-Spec}}
	\hfil
	\subfloat[]{\includegraphics[height=2.8in]{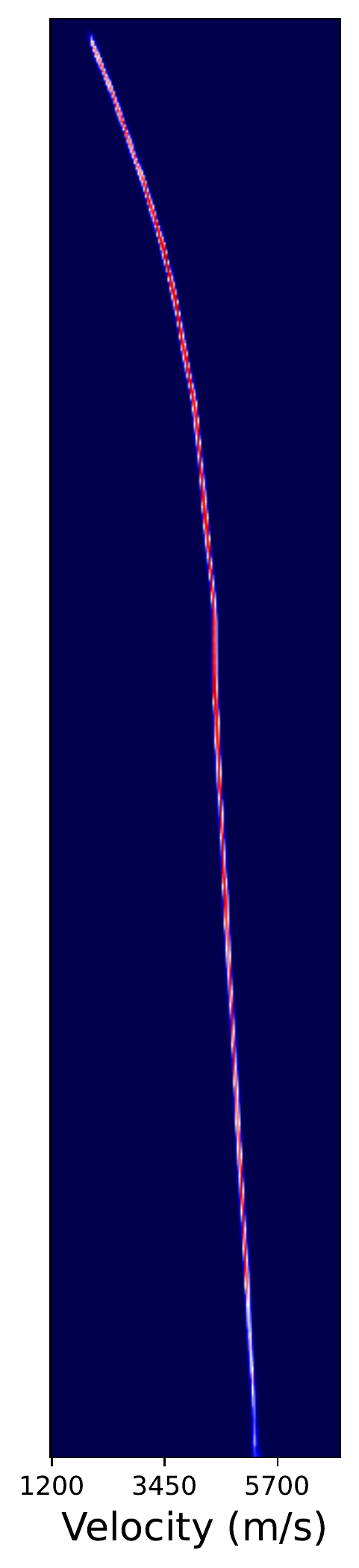}\label{fig: B-seg}}
	\hfil
	\subfloat[]{\includegraphics[height=2.8in]{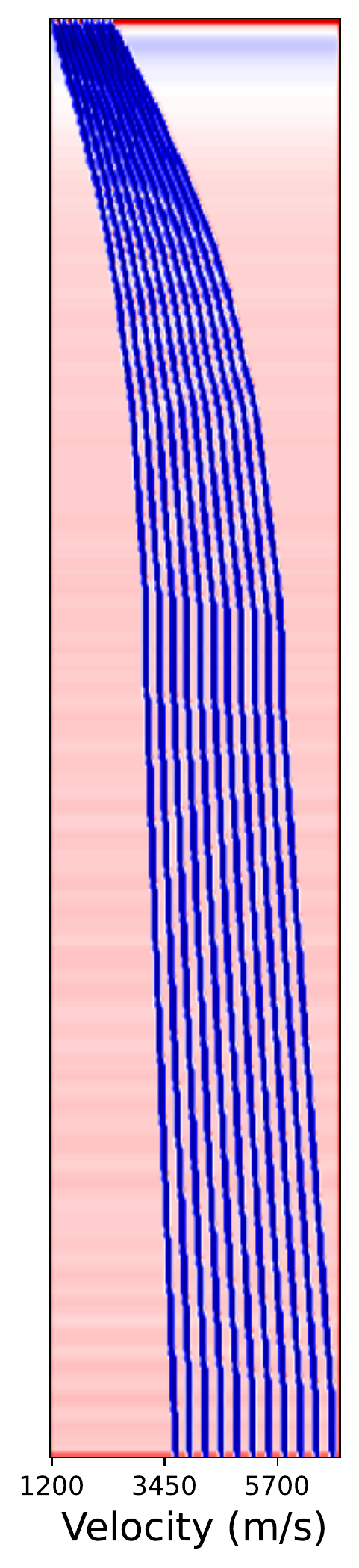}\label{fig: B-SGS}}
	\hfil
	\subfloat[]{\includegraphics[height=2.8in]{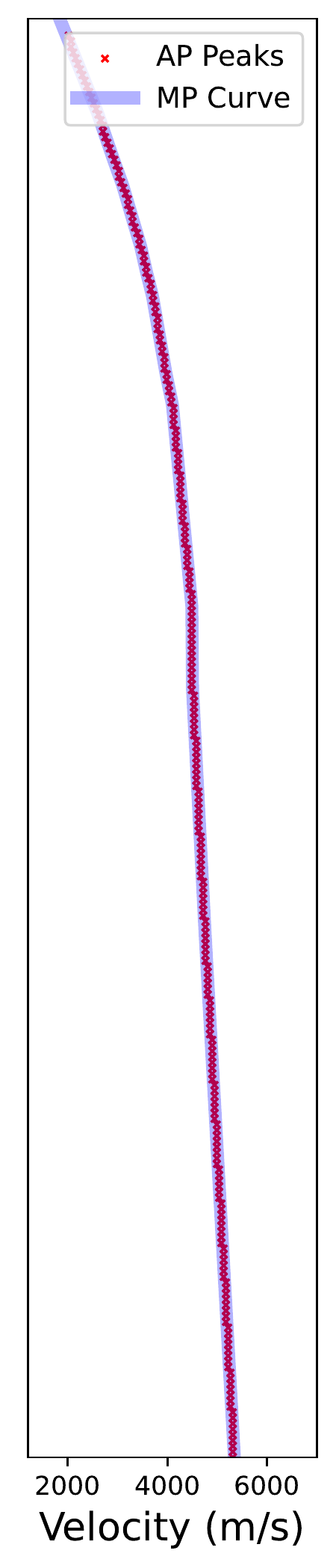}\label{fig: B-Vel}}
	\hfil
	\subfloat[]{\includegraphics[height=2.8in]{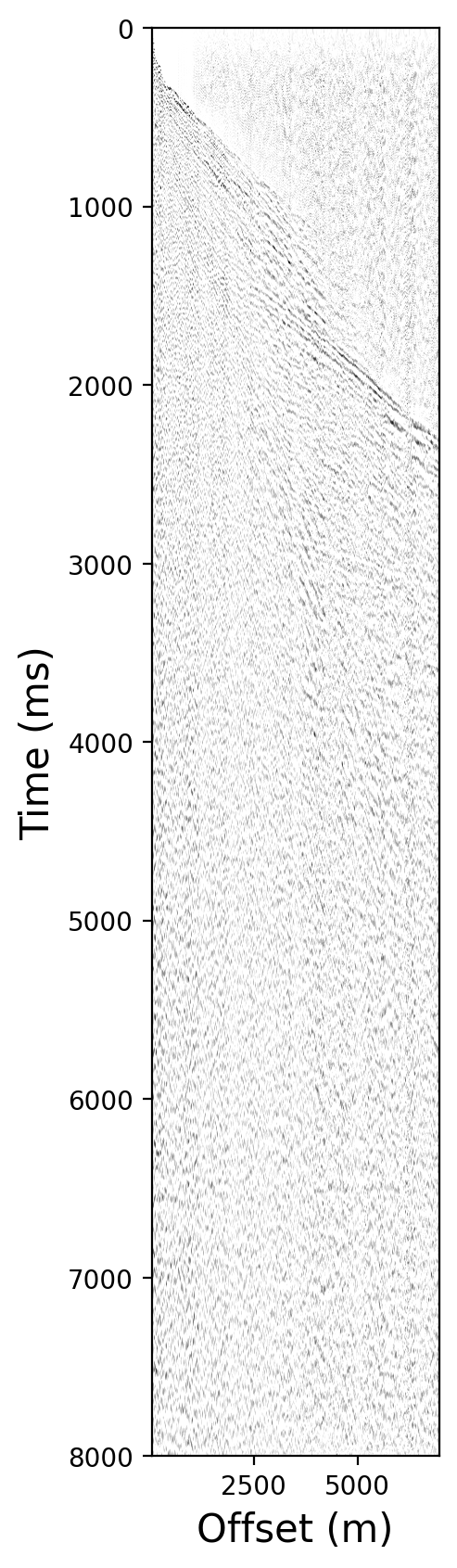}\label{fig: B-CMP}}
	\hfil
	\subfloat[]{\includegraphics[height=2.8in]{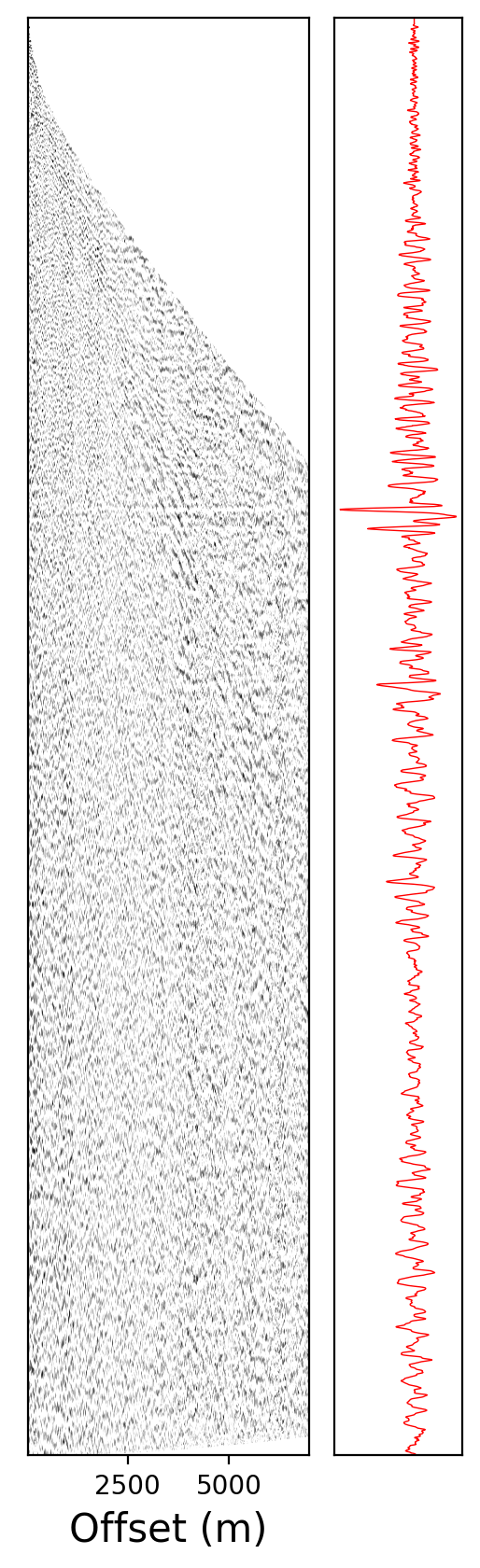}\label{fig: B-NMO}}
	\hfil
	\caption{The automatic picking result of MIFN. (a)-(f), (g)-(l) are the original velocity spectrum, segmentation map, feature map of SGS encoder, velocity curve (AV and MV), CMP gather and NMO CMP gather of data sets A (CMP line 2240 cdp 1840) and B (CMP line 940 cdp 2150) respectively. The right parts of (f) and (l) are the average values of each row.}
	\label{fig: SpecResult}
\end{figure*}

\begin{figure*}[ht!]
	\centering
	\subfloat[]{\includegraphics[width=0.16\linewidth]{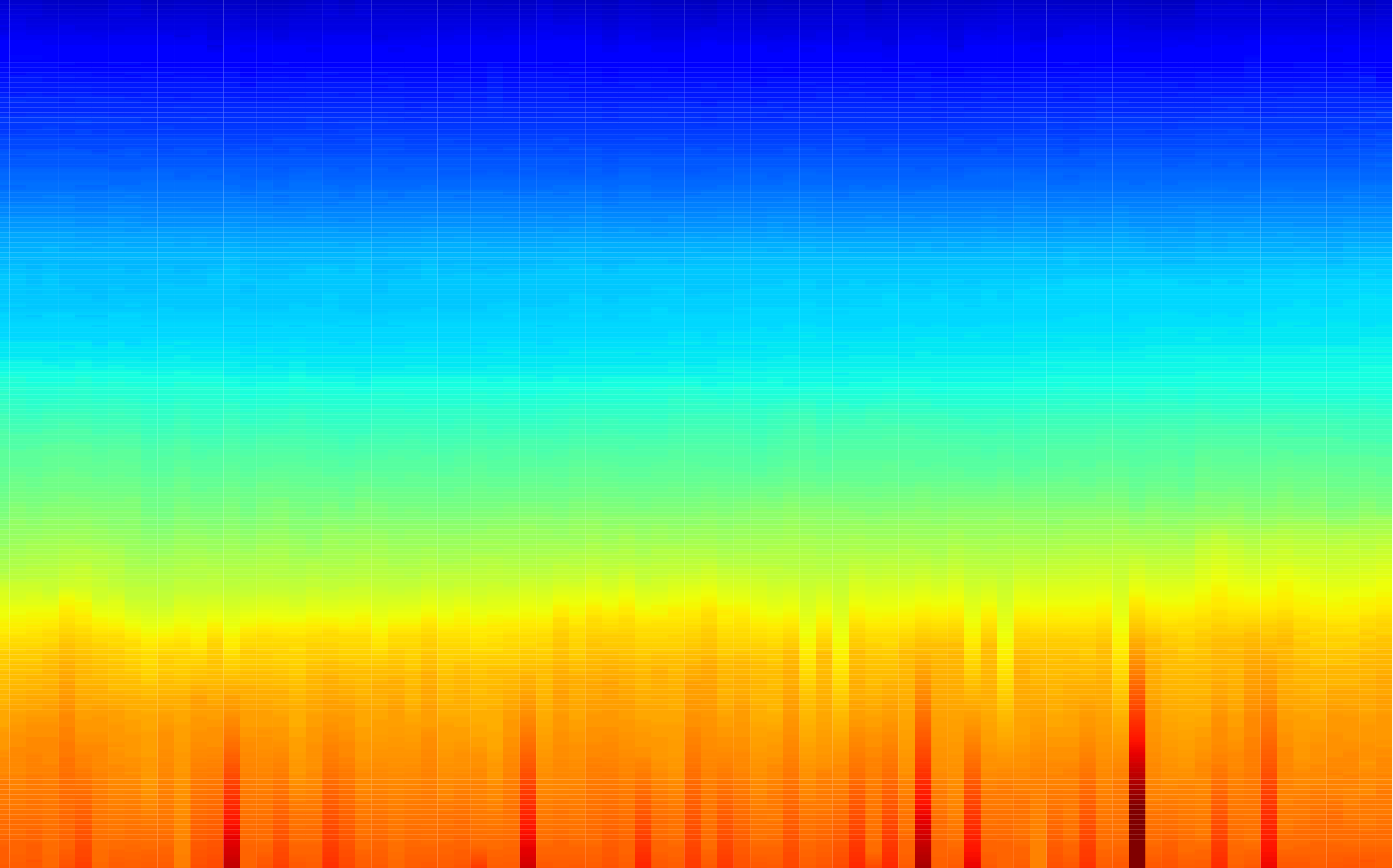}\label{fig: VF-SR-A-0.1}}\quad
	\subfloat[]{\includegraphics[width=0.16\linewidth]{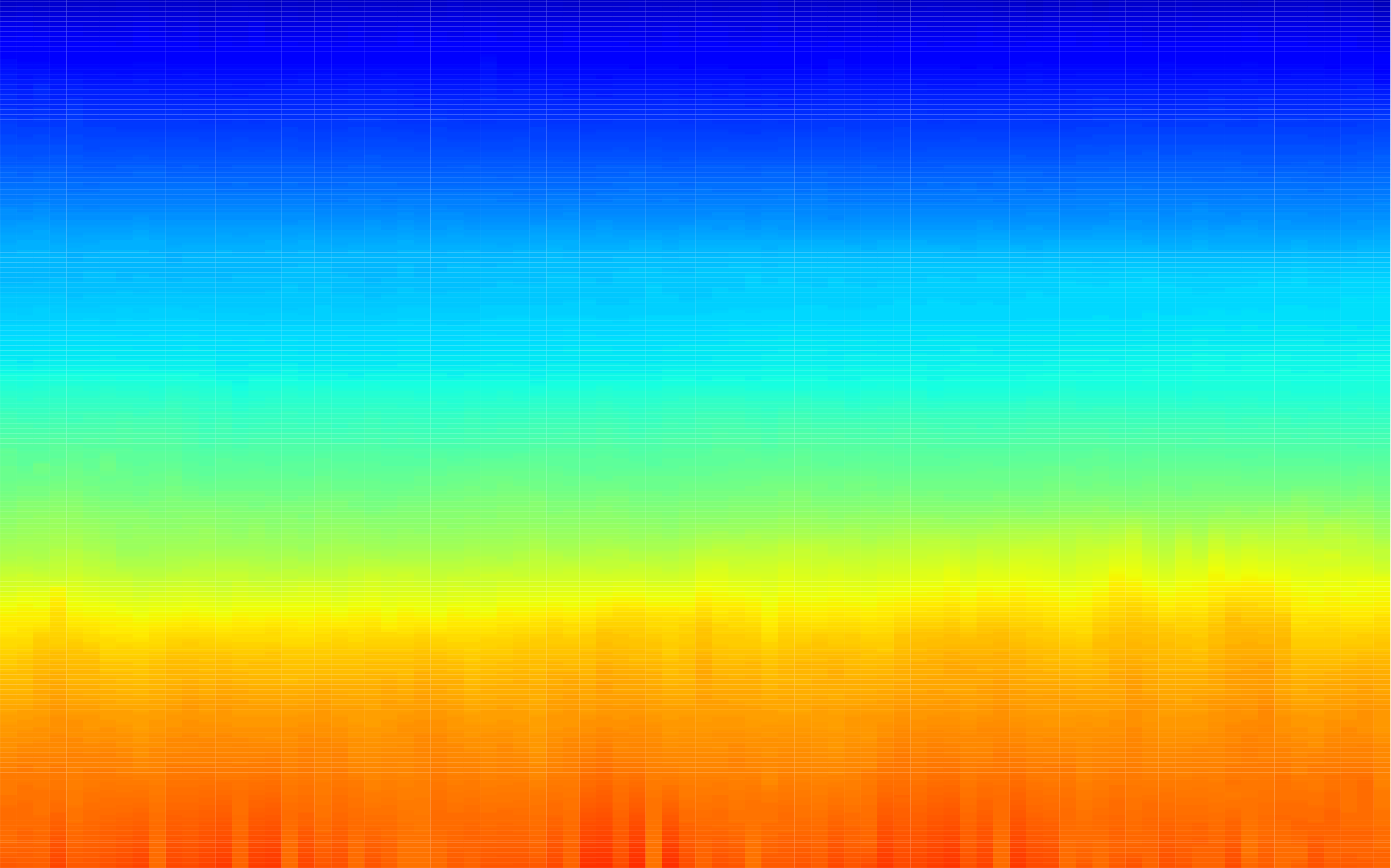}\label{fig: VF-SR-A-0.5}}\quad
	\subfloat[]{\includegraphics[width=0.16\linewidth]{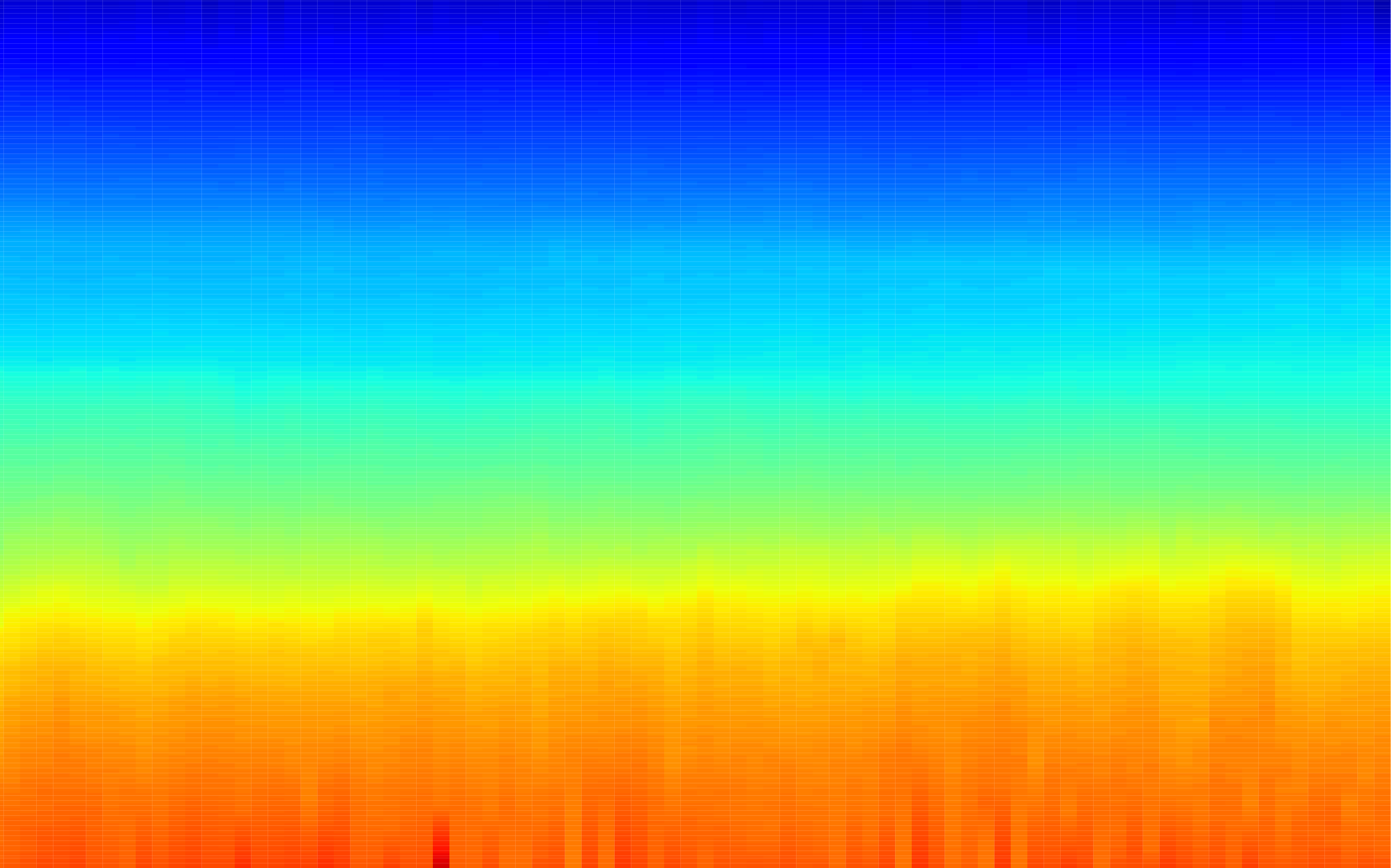}\label{fig: VF-SR-A-0.8}}\quad
	\subfloat[]{\includegraphics[width=0.16\linewidth]{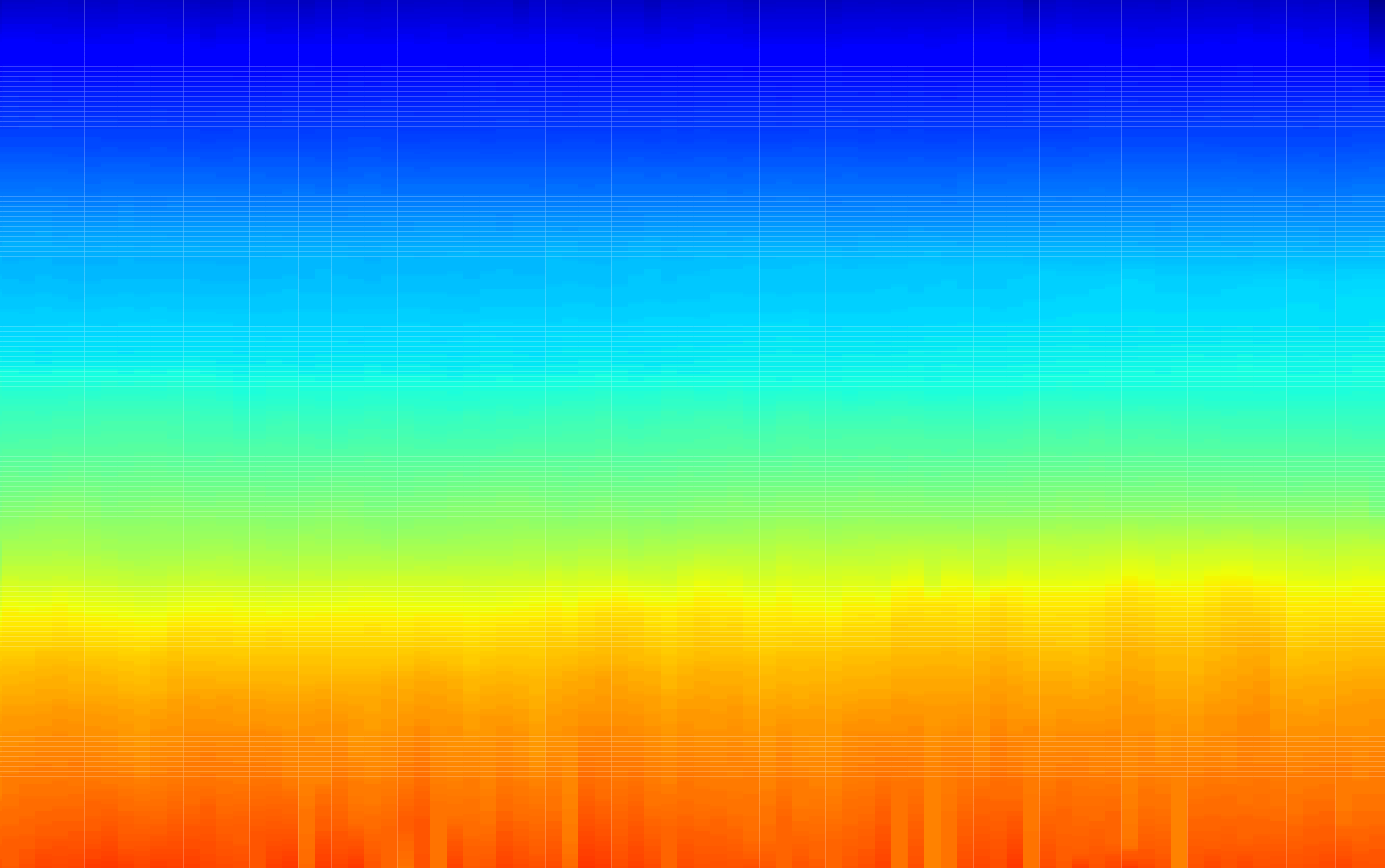}\label{fig: VF-SR-A-1}}\quad
	\subfloat[]{\includegraphics[width=0.173\linewidth]{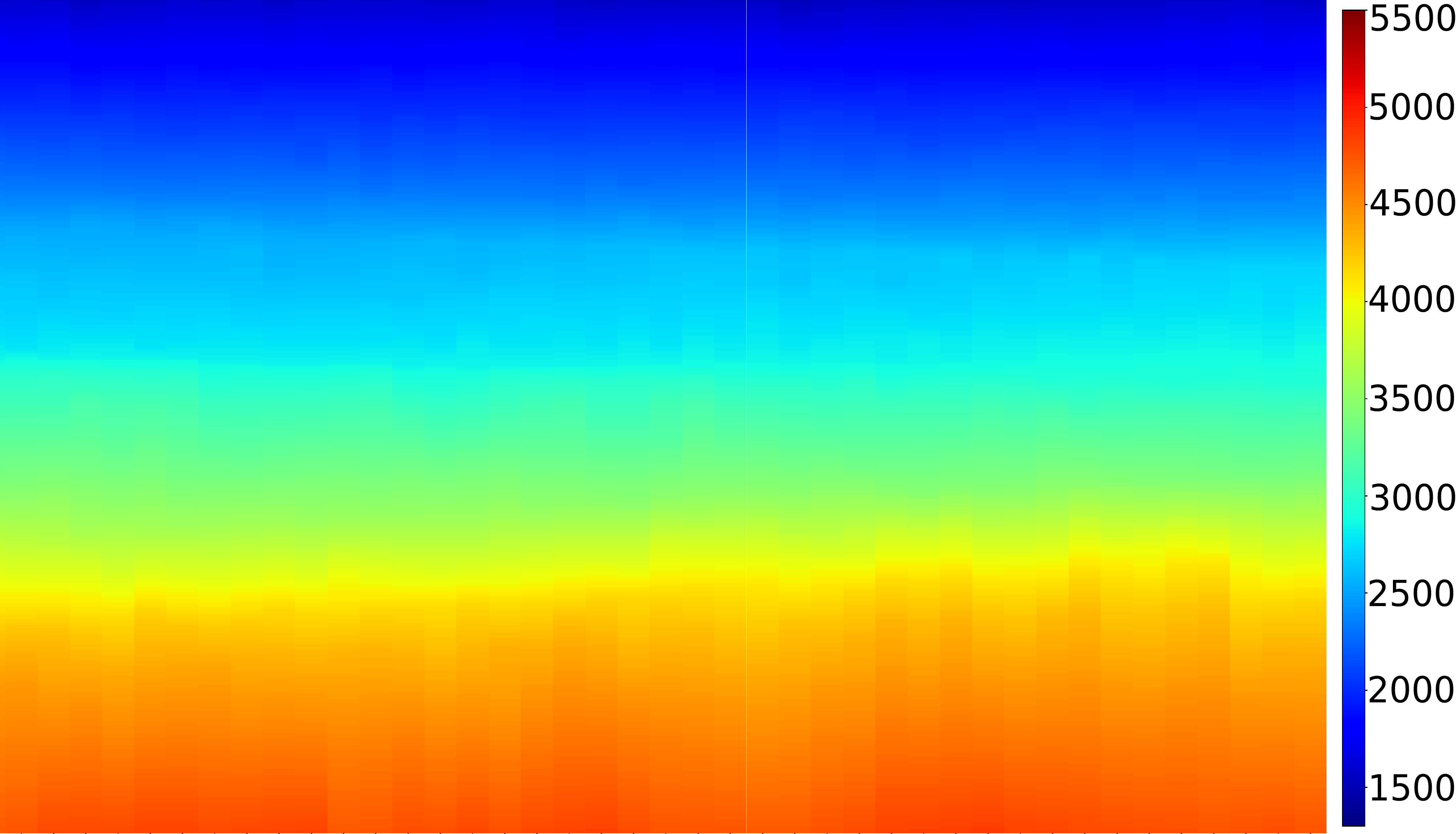}\label{fig: VF-A-MP}}
	\\
	\subfloat[]{\includegraphics[width=0.16\linewidth]{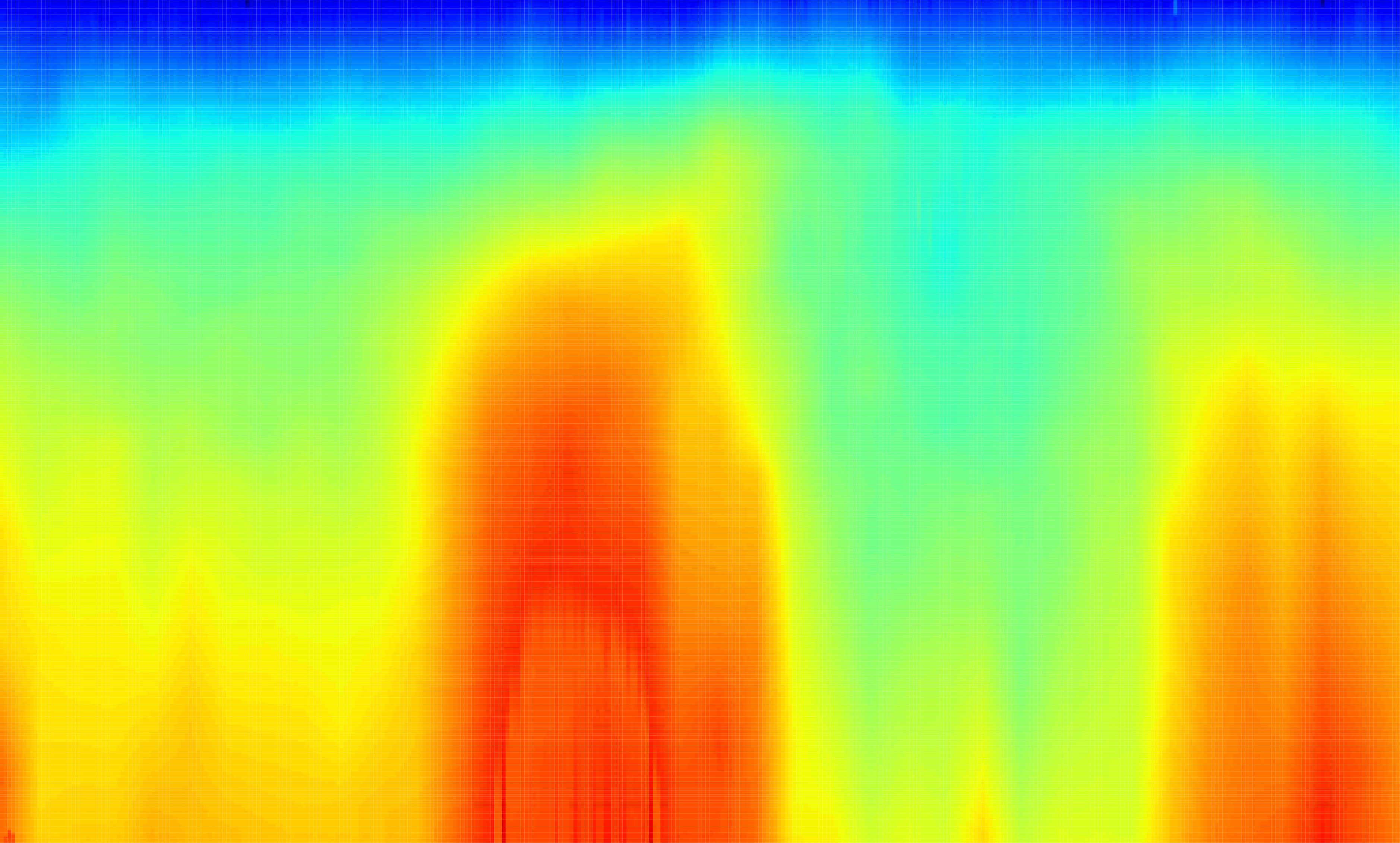}\label{fig: VF-SR-B-0.1}}\quad
	\subfloat[]{\includegraphics[width=0.16\linewidth]{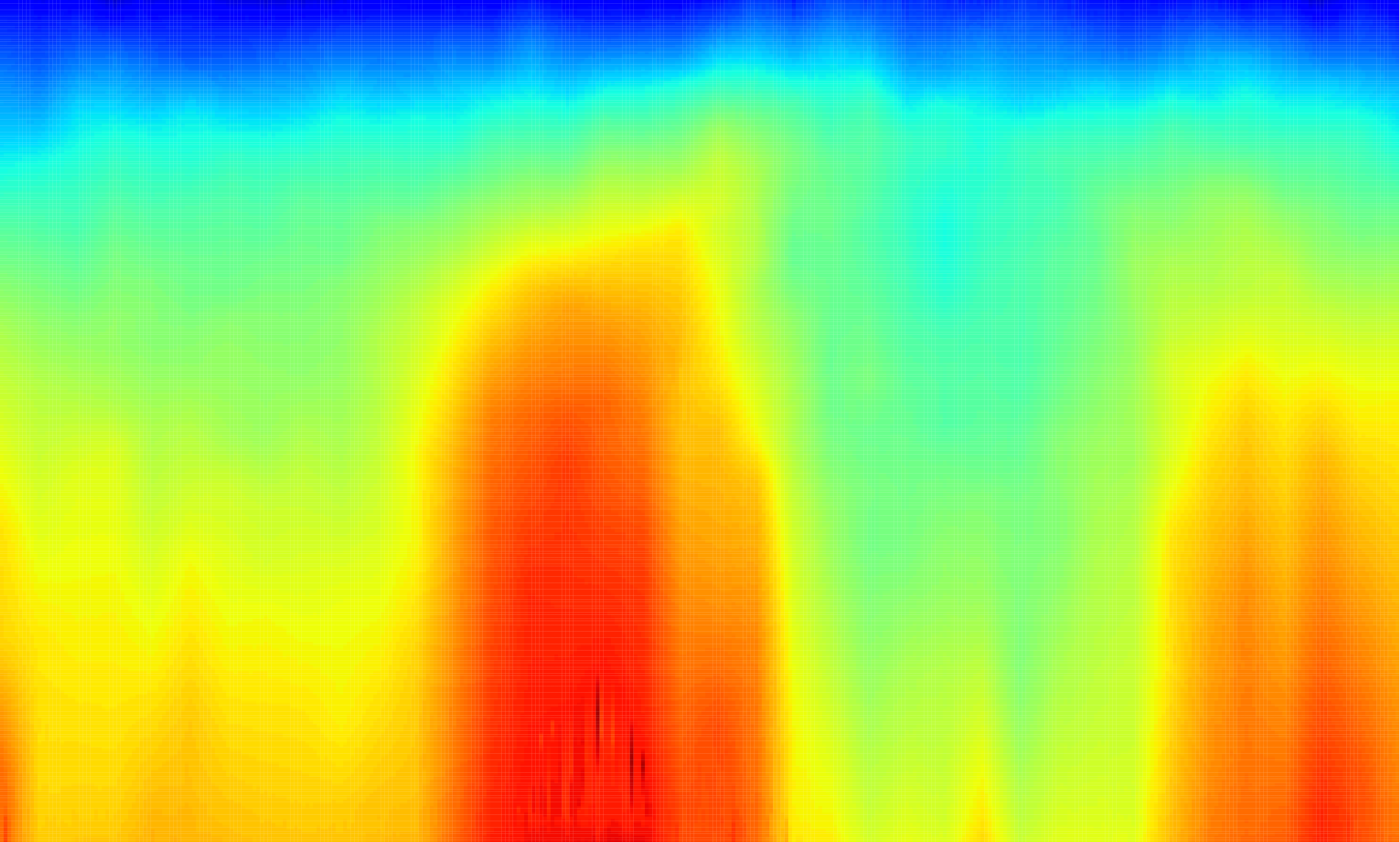}\label{fig: VF-SR-B-0.5}}\quad
	\subfloat[]{\includegraphics[width=0.16\linewidth]{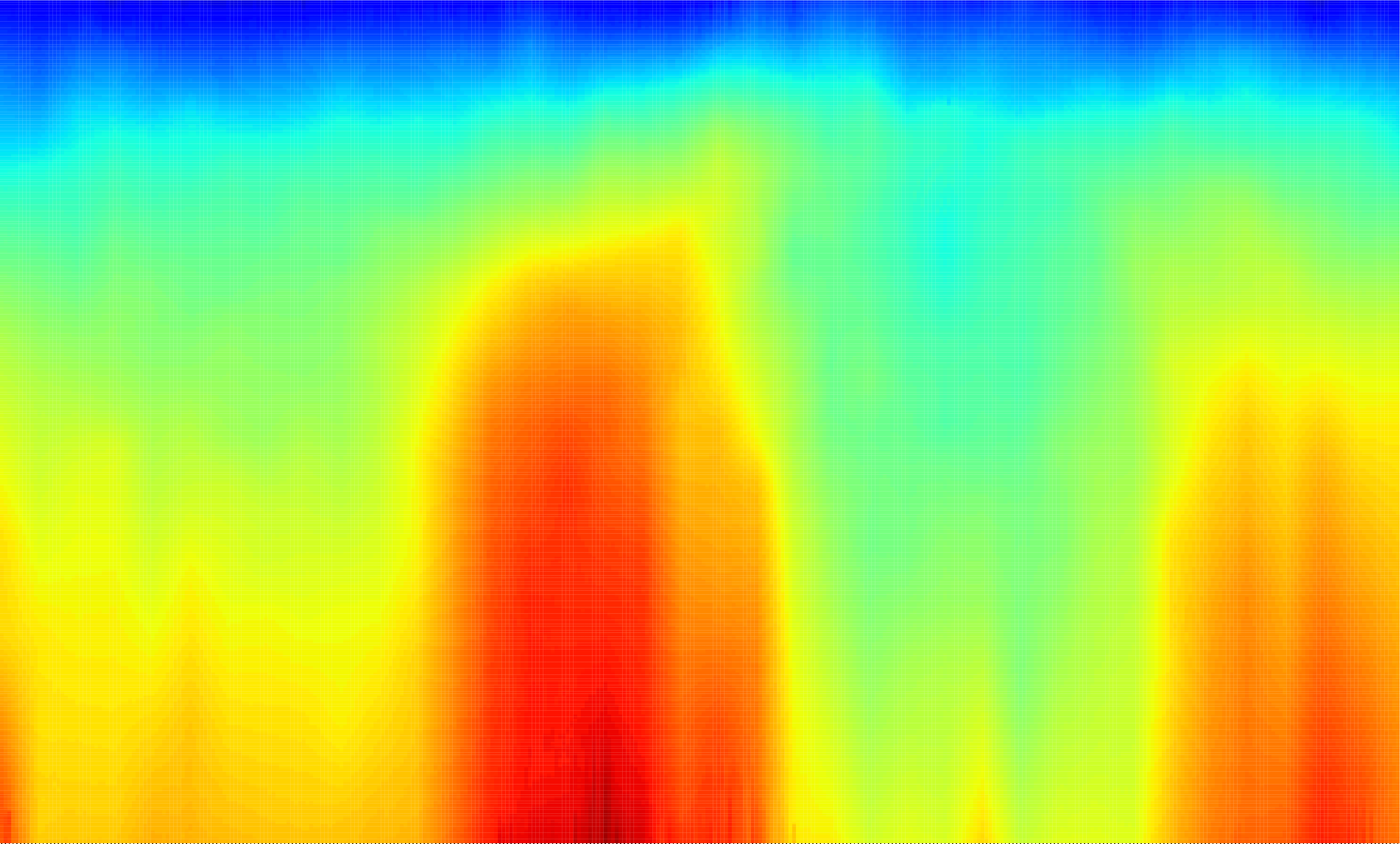}\label{fig: VF-SR-B-0.8}}\quad
	\subfloat[]{\includegraphics[width=0.16\linewidth]{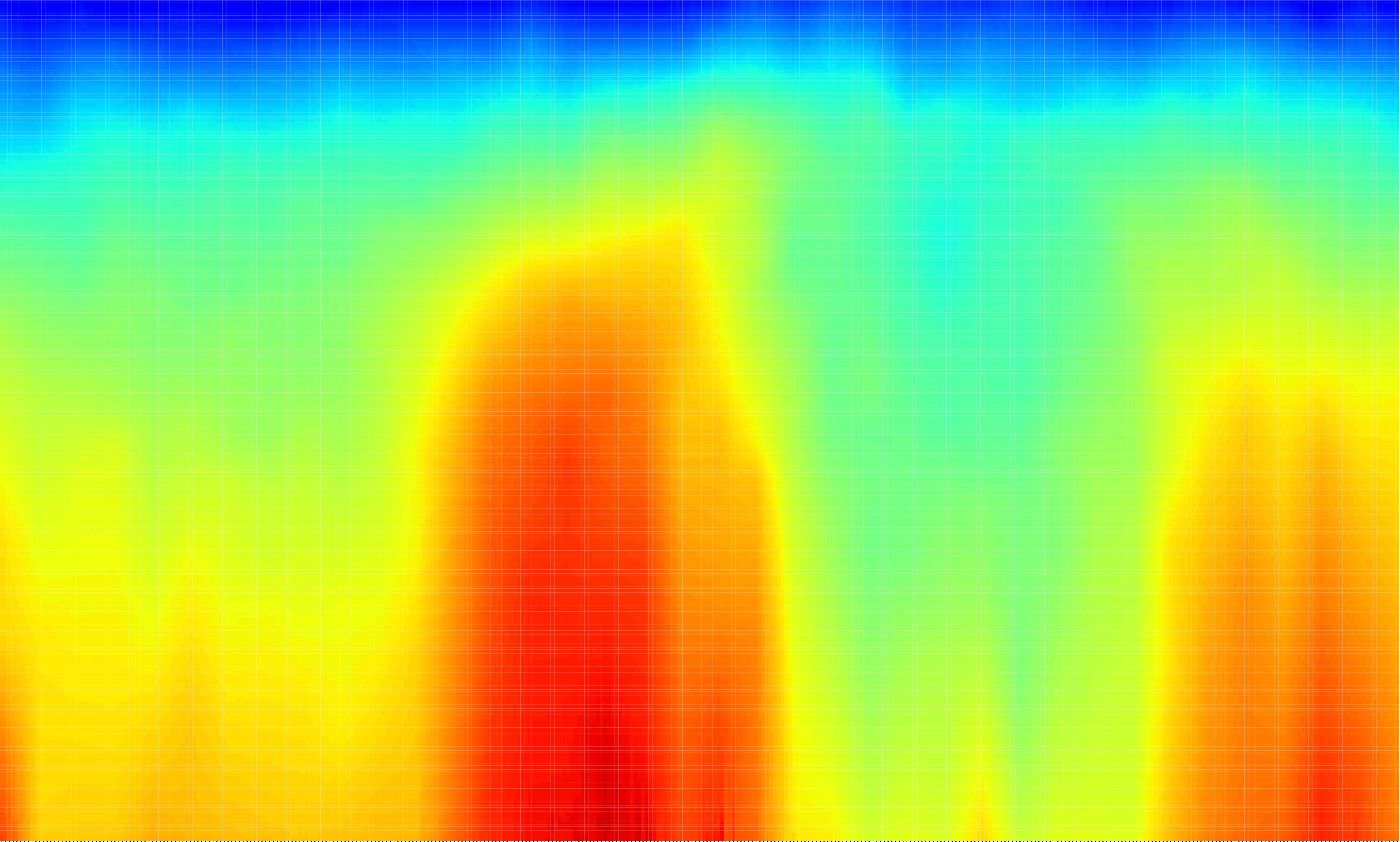}\label{fig: VF-SR-B-1}}\quad
	\subfloat[]{\includegraphics[width=0.174\linewidth]{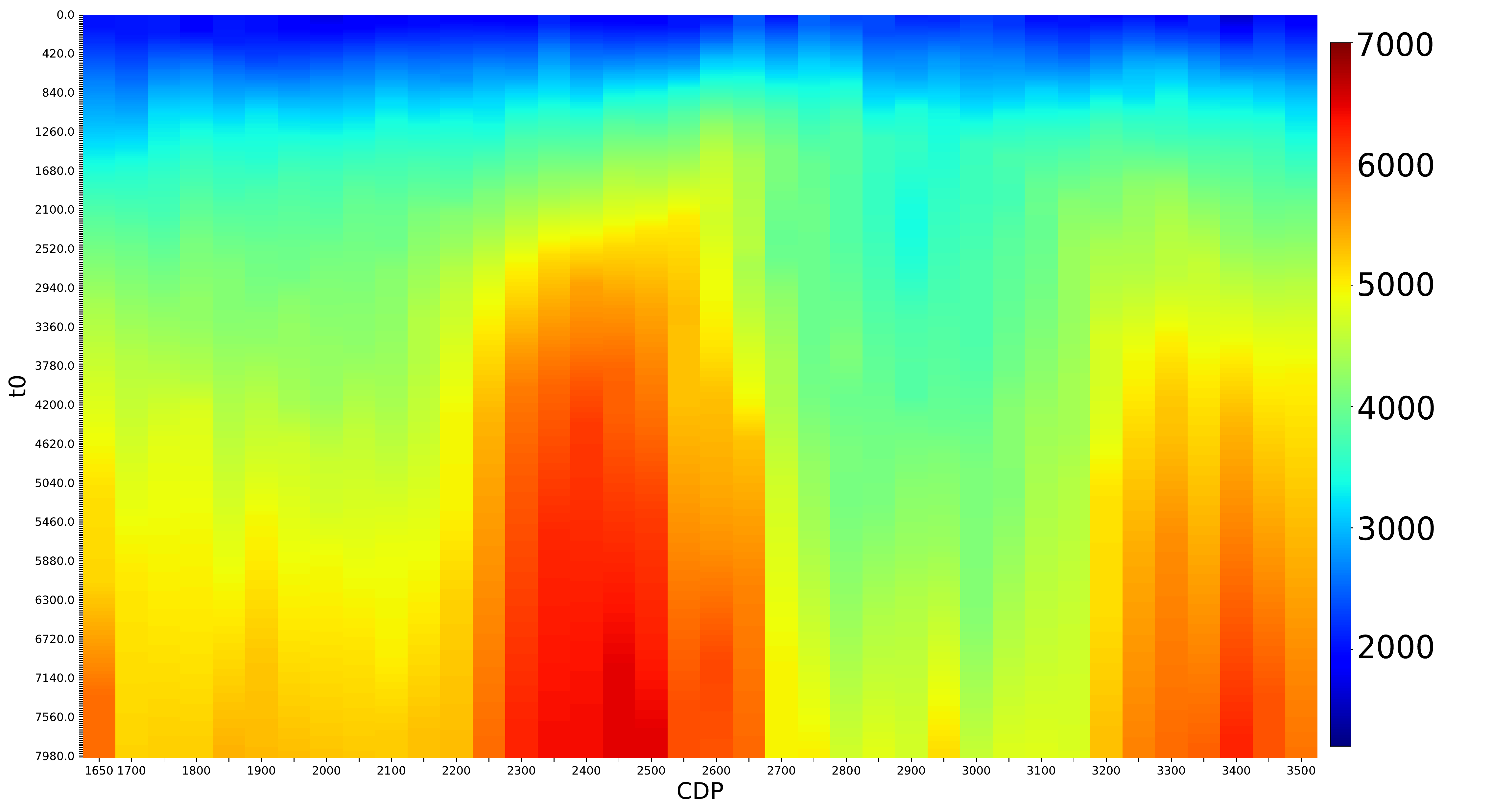}\label{fig: VF-B-MP}}
	\caption{Velocity field of stack velocity of picking results under different sampling rates. (a)-(d) and (f)-(i) are the MIFN picking results of line 3240 on dataset A and line 940 on dataset B, severally, under sampling rate = 0.2, 0.5, 0.8, 1. (e) and (j) are the manual picking results of line 3240 on dataset A and line 940 on dataset B, respectively.}
	\label{fig: VelField}
\end{figure*}

\subsubsection{Ablation Experiment}
We perform an ablation experiment in which we put up model variants with varied structures using the control variable method, to better understand the specific contribution of each component of our proposed model. We train all model on field datasets A and B, with the same training strategies as written in Section III.B, and then test these model on the same test set. In Tab. \ref{tab:ablation}, we study the different components of MIFN by removing the spectrum feature extractor (SFE), SGS encoder and the post-processing method. Specifically, removing SFR, denoted by w/o SFE, that is, the input of U-Net is changed to the velocity spectrum and the feature map obtained by the SGS encoder. Removing the SGS encoder, represented by w/o SGS encoder, and the input of U-Net is set as the original velocity spectrum as well as its 10 multi-scale observation feature maps, and the others remain unchanged. The last one is the w/o post-processing, which means that the model without our post-processing method uses linear extrapolation to estimate the shallow and deep velocities from the probability segmentation image, which is the output of U-Net. As shown in Tab. \ref{tab:ablation}, the results of the test set show that SFE, SGS encoder and our post-processing are very reasonable settings and can improve the quality of our picking results. 

\begin{table}[ht!]
\centering
\caption{ABLATION EXPERIMENT ON THE TEST SET}
\label{tab:ablation}
\begin{tabular}{lcc}
\toprule\toprule
Model Variants      & VMAE of Dataset A & VMAE of Dataset B \\ \midrule
w/o SFE             & 25.395  & 15.634  \\
w/o SGS encoder     & 26.743  & 124.656  \\
w/o post-processing & 21.594  & 13.423  \\ \hline
Ours                & \textbf{17.209}  &  \textbf{12.798} \\ \bottomrule\bottomrule
\end{tabular}
\end{table}

\subsubsection{Generalization Ability Test}
We design another experiment named generalization ability test to verify the generalization ability of our model trained by different numbers of training samples. As Fig. \ref{fig: dist-A} shown, we split the original train set to two parts denoted by train set and unused set, using a series sampling percentage, which are recorded as the sampling rates. To keep the same test condition, we do not change the validation set and test set in this experiment. There are four models trained with different sampling rates on data sets A and B, respectively, and the training strategy is the same as section III.B. In Fig. \ref{fig: LA} and \ref{fig: LB}, the BCE losses of the validation set and training set decrease smoothly and converges, and there is no over-fitting phenomenon, which verifies the feasibility of our model. 
As Fig. \ref{fig: ABAll} shown, VMAE on the test set increases, when the number of training samples decreases. 
When the sampling rate = 0.2, the sample numbers for training of data sets A and B are 96 and 136, respectively, in which it is already a small sample learning environment in deep learning. But the VMAE can still be 28.38 m/s and 23.72 m/s on data sets A and B.
In order to observe the differences at different sampling rates more macroscopically, we plot the velocity field of the picked results at sampling rates equal to 0.2, 0.5, 0.8, 1, as shown in Fig. \ref{fig: VelField}. 
It shows that when the sampling rate = 0.2, a relatively rough stack velocity model can be picked up, and with the increase of the number of the sampling rate, the picking accuracy is improved. But when the sampling rate is 0.8, the accuracy is really close to the accuracy of sampling rate = 1. That means, when the number of samples reaches a sufficient number, the improvement is is not obvious enough,  as Fig. \ref{fig: ABAll} and Fig. \ref{fig: VelField}
shown. This result implies that in the actual seismic field data processing, the cost of manual labeling can be reduced at the expense of small precision. 

\subsubsection{Performance of MIFN}  
After verifying the accuracy and feasibility of the model, there are some visual details of the velocity pickup of MIFN. In general, the automatically picked velocity curve (AV) on data set A or B is very close to the manually picked velocity curve (MV), as shown in Fig. \ref{fig: A-Vel} and \ref{fig: B-Vel}. Even at key turning points there are good pickup results.

The medium SNR velocity spectrum (dataset A) contains enough semantic information for velocity picking. As shown in Fig. \ref{fig: A-Spec} and \ref{fig: A-Vel}, the trend of segmentation is along the direction of the energy clusters. 
However, for the velocity spectrum with low SNR (dataset A, Fig. \ref{fig: B-Spec}), there is not enough information to pick velocity. We plot the output feature map of SGS encoder in MIFN as Fig. \ref{fig: B-SGS} shown, in which we explore how SGS information assists the velocity picking. It can be observed from the Fig. \ref{fig: A-SGS} and \ref{fig: B-SGS} that the feature map contains two parts of information. One is the stack velocity prior, which comes from the scanning stack velocity curve. The other is the weights of these velocity prior, which are represented by the value on the feature map. In Fig. \ref{fig: B-seg}, segmentation result is depended on the SGS information, and the curves in Fig. \ref{fig: B-SGS} are as a prior information to guide the picking. 

\begin{figure}[ht!]
	\centering
	\subfloat[]{\includegraphics[width=1.6in]{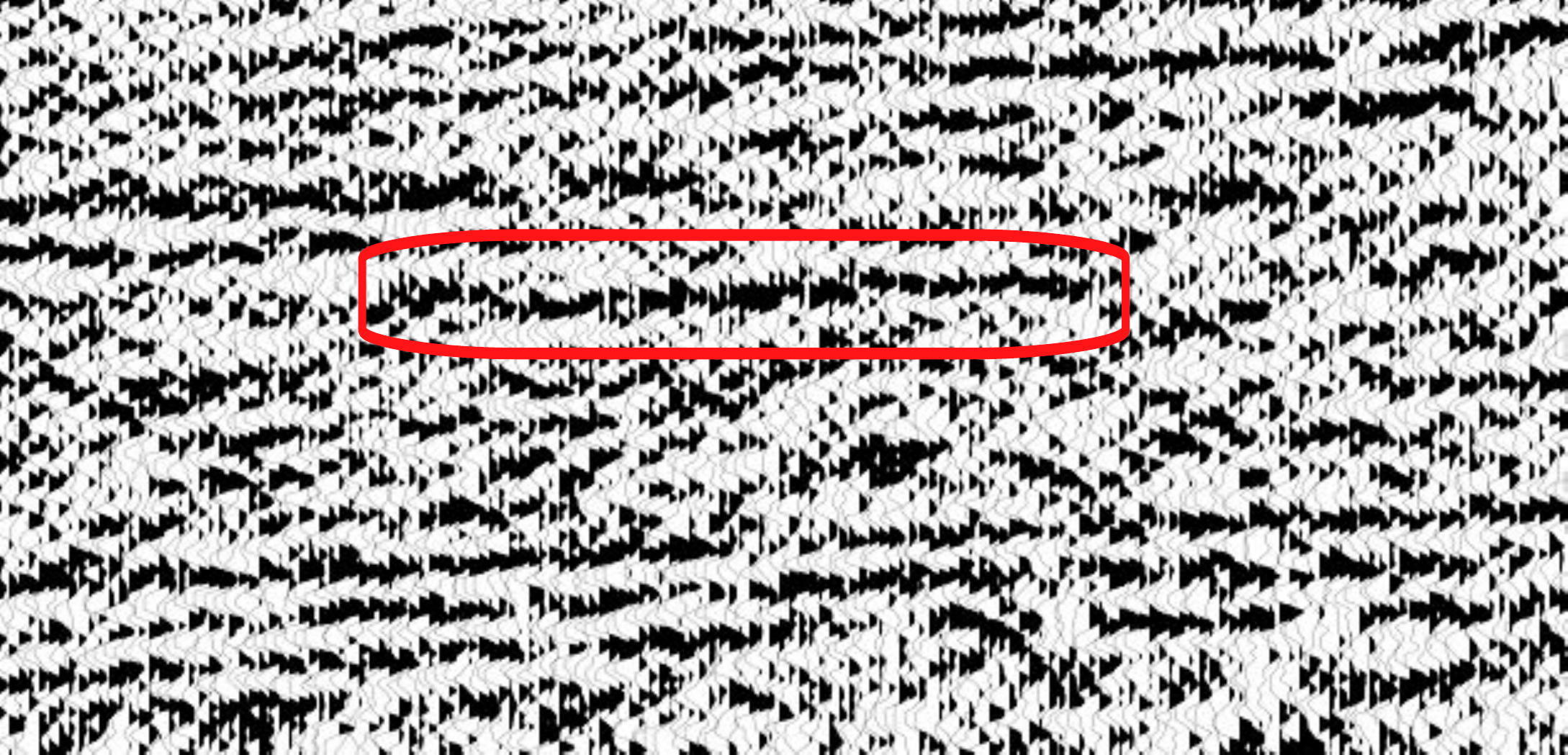}\label{fig: stk-1-A}}\quad
	\subfloat[]{\includegraphics[width=1.6in]{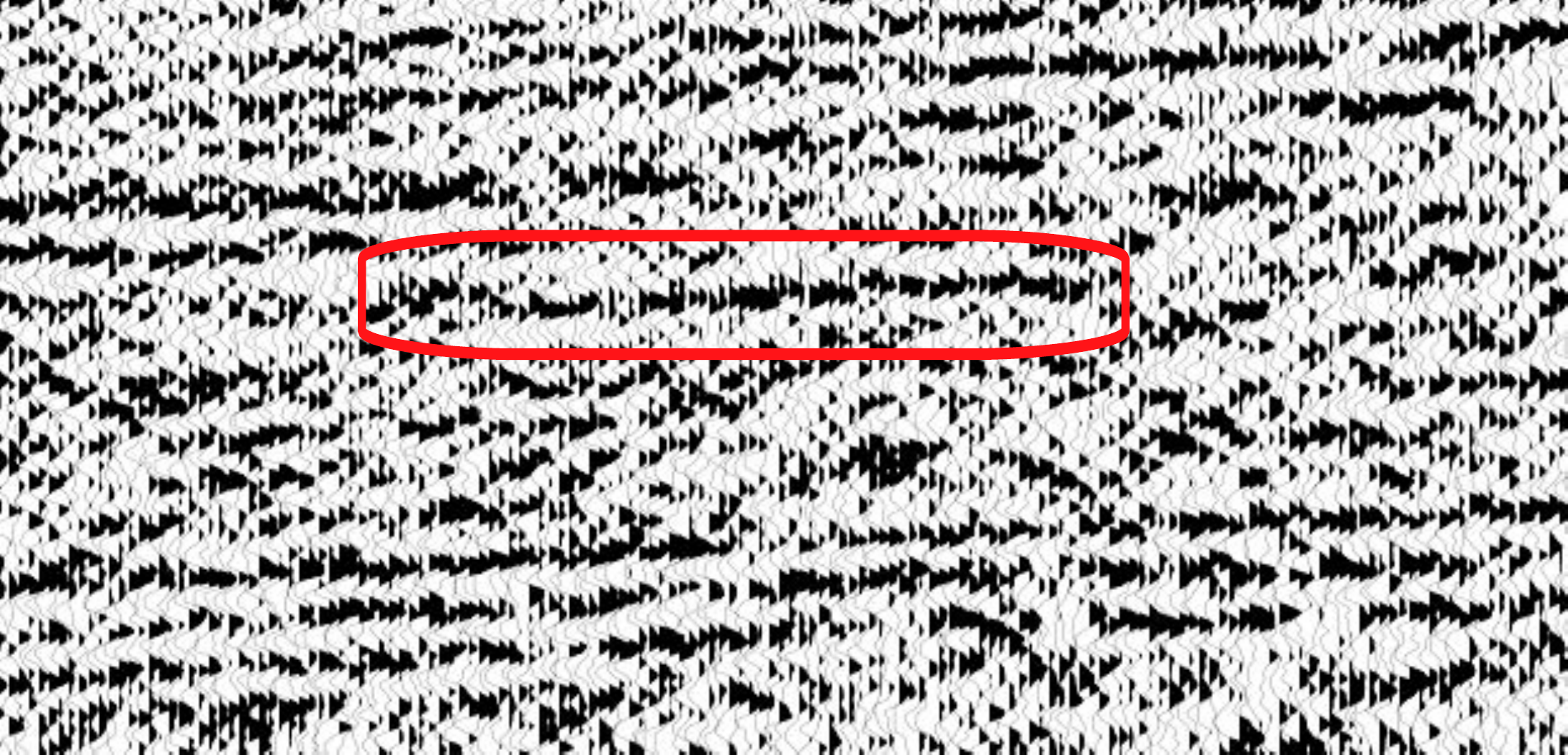}\label{fig: stk-1-M}}\\
	\subfloat[]{\includegraphics[width=1.6in]{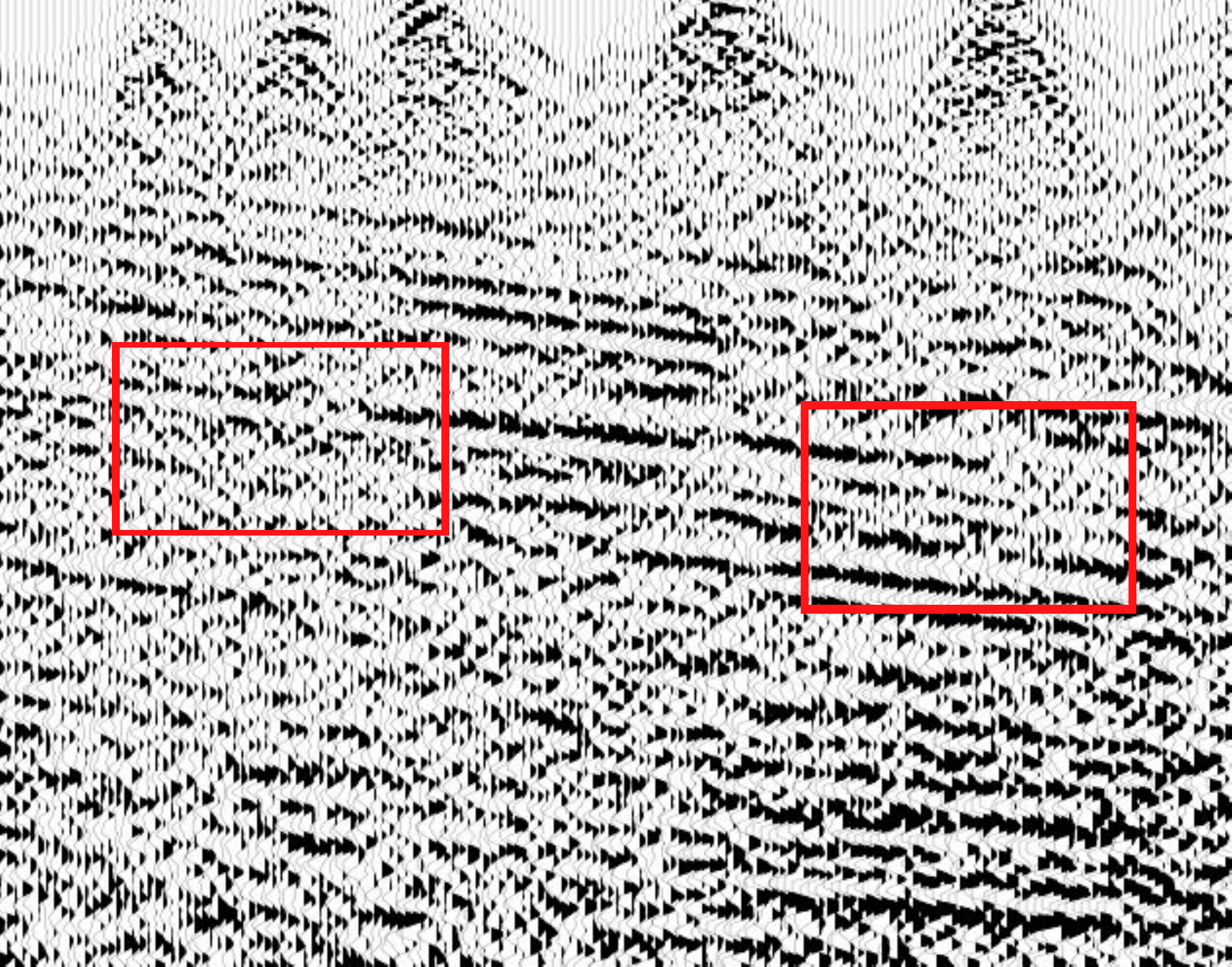}\label{fig: stk-2-A}}\quad
	\subfloat[]{\includegraphics[width=1.6in]{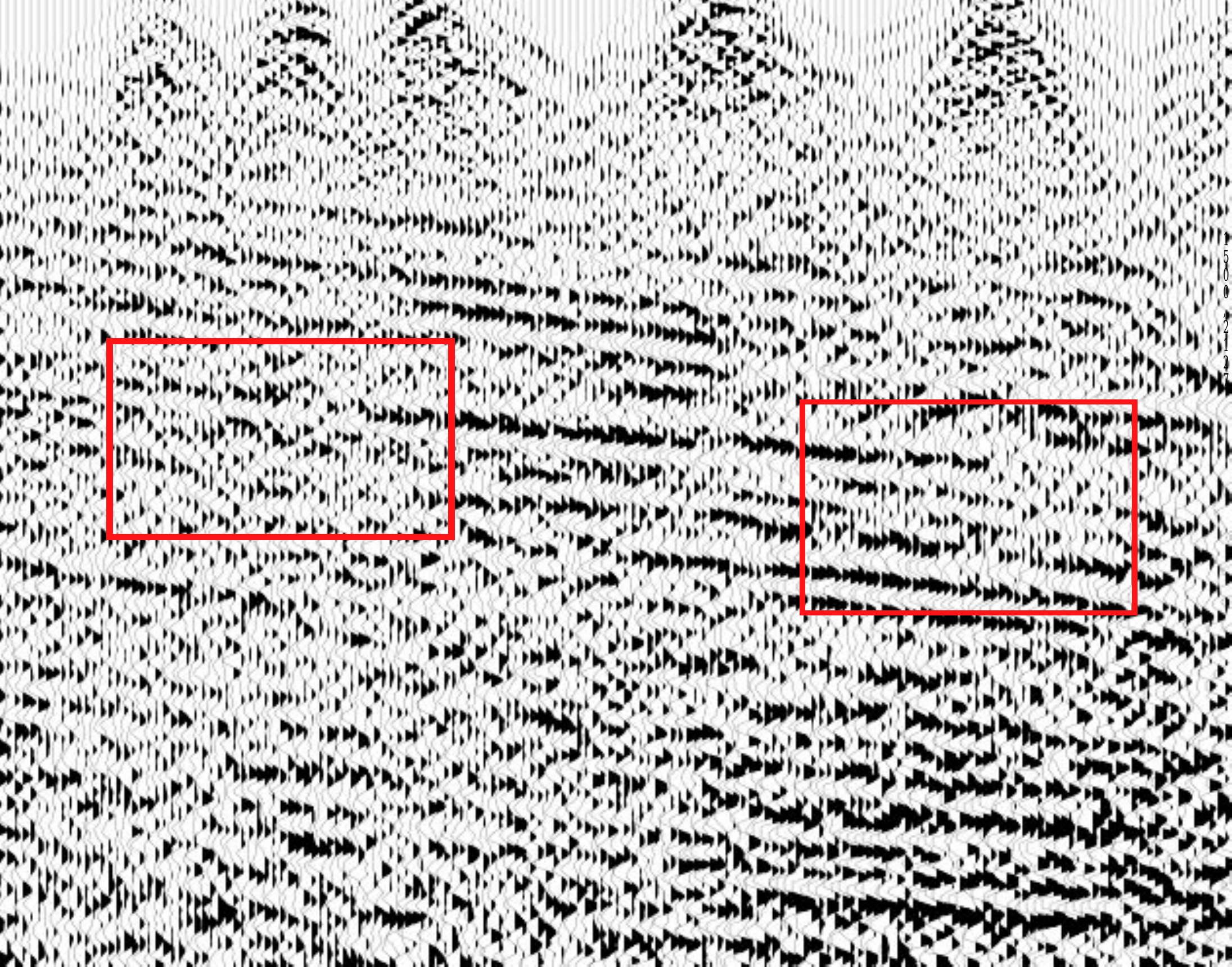}\label{fig: stk-2-M}}
	\caption{Some details of the stacked NMO CMP gather on dataset B (line 940). (a) , (c) are the automatic results, and (b), (d) is the corresponding manual result}
	\label{fig: stk}
\end{figure}

Then, we use the stacking velocity which MIFN picks to generate stacked NMO CMP gathers and velocity field images. In Fig. \ref{fig: stk}, compared with the manual picking results (Fig. \ref{fig: stk-1-M} and \ref{fig: stk-2-M}), the automatic picking results (Fig. \ref{fig: stk-1-A} and Fig. \ref{fig: stk-2-A}) have better continuity and the texture is more obvious. As shown in Fig. \ref{fig: VF-A-MP} and Fig. \ref{fig: VF-B-MP}, the stack velocity of manual annotation is sparse, and the auto-picking velocity of MIFN is very close to the manual picking result in Fig. \ref{fig: VF-SR-A-1} and Fig. \ref{fig: VF-SR-B-1}.

\subsubsection{Fine-tuning Model Test}
To evaluate the transfer ability and the practicality of MIFN, we adopt a fine-tuning-based few-shot learning approach to train a fine-tuning model, in which we train a pretrained model in advance on dataset A and B using all train samples, respectively, and fine-tune them on a new field dataset using partial train samples. Specifically, we select three sampling rates (SR), i.e. 0, 0.2, 0.5. All other training strategies are consistent with Section III.B. Unfortunately, direct transfer of picking results on other models is suboptimal, where the VMAEs are 156.59 m/s and 485.74 m/s, respectively. As shown in Tab. \ref{tab: fine-tune}, the fine-tuned model has a 38.51\% and 27.02\% improvement in accuracy under SR=0.2 on datasets A and B, respectively, compared to the directly trained model. We analyze that the promotion may be attributed to a lot of prior information is preserved in the pretrained model, which is exactly what few-shot learning expects. Moreover, with the increase of training sample number, the influence of prior information in the pretrained model decreases, so the improvement of picking accuracy is not so obvious compared with SR=0.2, and the improvement of A and B is 9.68\% and 13.06\% in the case of SR=0.5.
Through the above results, we confirm the transfer ability and application ability of our model in industrial production.

\begin{table}[ht!]
\centering
\caption{Fine-tuning model test VMAE results}
\label{tab: fine-tune}
\begin{tabular}{ccccc}
\toprule\toprule
\multirow{2}{*}{\begin{tabular}[c]{@{}c@{}}Test\\ Data\end{tabular}} &
  \multirow{2}{*}{\begin{tabular}[c]{@{}c@{}}Fine-tuning\\ Data\end{tabular}} &
  \multicolumn{3}{c}{Mean VMAE(m/s)} \\ \cline{3-5} 
 &
   &
  No Pretraining &
  \begin{tabular}[c]{@{}c@{}}Pretrained\\ Model\end{tabular} &
  Improvement \\ \hline
\multirow{3}{*}{A} & -         & - & 156.59 & - \\
                   & A(SR=0.2) & 28.38  & 17.45 & 38.51\%  \\
                   & A(SR=0.5) & 18.70  & 16.89 & 9.68\%  \\ \hline
\multirow{3}{*}{B} & -         & - & 485.74 & - \\
                   & B(SR=0.2) & 23.72  & 17.31 & 27.02\%  \\
                   & B(SR=0.5) & 15.62  & 13.58 & 13.06\%  \\ \bottomrule\bottomrule
\end{tabular}
\end{table}

\section{Conclusion}
We assume that the automatic velocity spectrum picking problem is a semantic segmentation task. We have developed a deep learning model named MIFN, which is based on U-Net to solve this task. In our work, we expect to improve the picking accuracy of the spectrum with low SNR. Thus, we propose the velocity spectrum enhance method and the fusion method of multi-information. 

We evaluated MIFN on two field datasets, i.e., low SNR and medium SNR. The results indicates that the semantic segmentation is suitable for velocity picking problem, and MIFN can pick the spectrum with low SNR well. Specifically, the mean deviation is lower than 20 m/s. The addition of the SGS can improve the picking accuracy dramatically, especially in the case of low SNR. We also try a few-shot method to train a pretrained model and fine-tune on a new area, and the results show that fine-tuned MIFN can achieve good performance, but the non-fine-tuned model cannot generalize well on the new dataset. In the future work, we will improve the SGS encoder into an encoder with adaptive input width to fit the input SGS of different widths. To improve the generalization, we will also consider training our method on more data sets and forming a big pretrained model, and conduct more few-shot learning methods. 




\ifCLASSOPTIONcaptionsoff
  \newpage
\fi

\bibliographystyle{IEEEtran}
\bibliography{reference}

\begin{IEEEbiography}[{\includegraphics[width=0.9in,height=1.25in,clip]{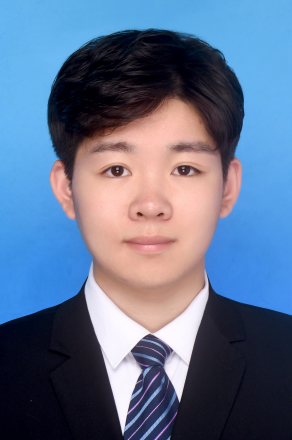}}]{Hongtao~Wang} is currently pursuing the Ph.D. degree in statistics with the School of Mathematics and Statistics, Xi’an Jiaotong University, Xi’an, China. His research interests include Bayesian statistics and deep learning.
\end{IEEEbiography}

\begin{IEEEbiography}[{\includegraphics[width=0.9in,height=1.25in,clip]{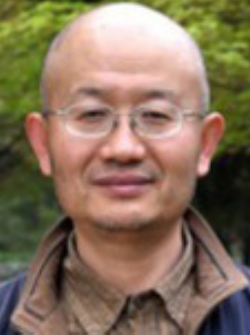}}]{Jiangshe~Zhang} was born in 1962. He received the M.S. and Ph.D. degrees in applied mathematics from Xi'an Jiaotong University, Xi'an, China, in 1987 and 1993, respectively, where he is currently a Professor with the Department of Statistics. He has authored and co-authored one monograph and over 80 conference and journal publications on robust clustering, optimization, short-term load forecasting for the electric power system, and remote sensing image processing. His current research interests include Bayesian statistics, global optimization, ensemble learning, and deep learning.
\end{IEEEbiography}

\begin{IEEEbiography}[{\includegraphics[width=0.9in,height=1.25in,clip]{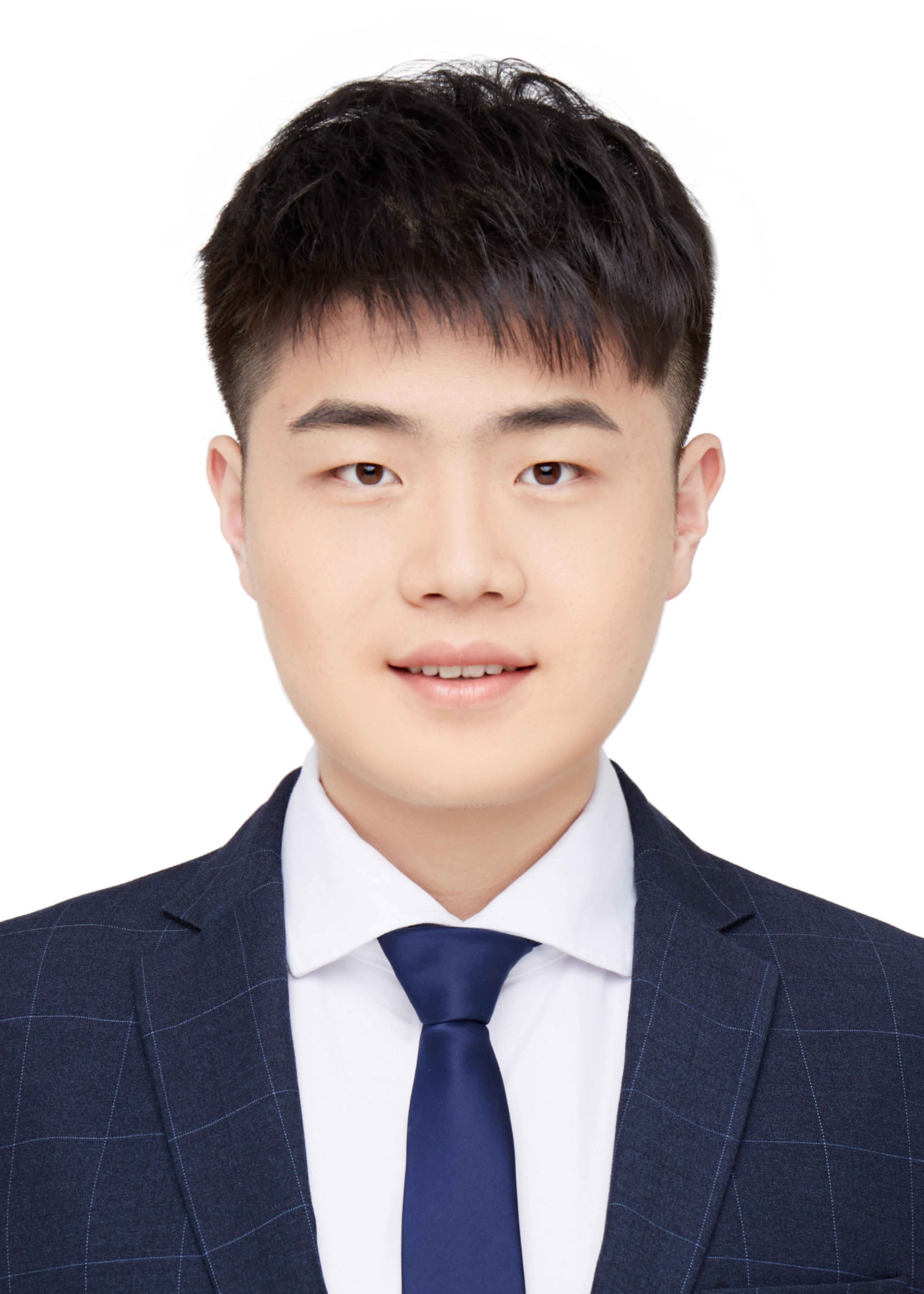}}]{Zixiang~Zhao} is currently pursuing the Ph.D. degree in statistics with the School of Mathematics and Statistics, Xi’an Jiaotong University, Xi’an, China. His research interests include computer vision, deep learning and low-level vision.\end{IEEEbiography}

\begin{IEEEbiography}[{\includegraphics[width=0.9in,height=1.25in,clip]{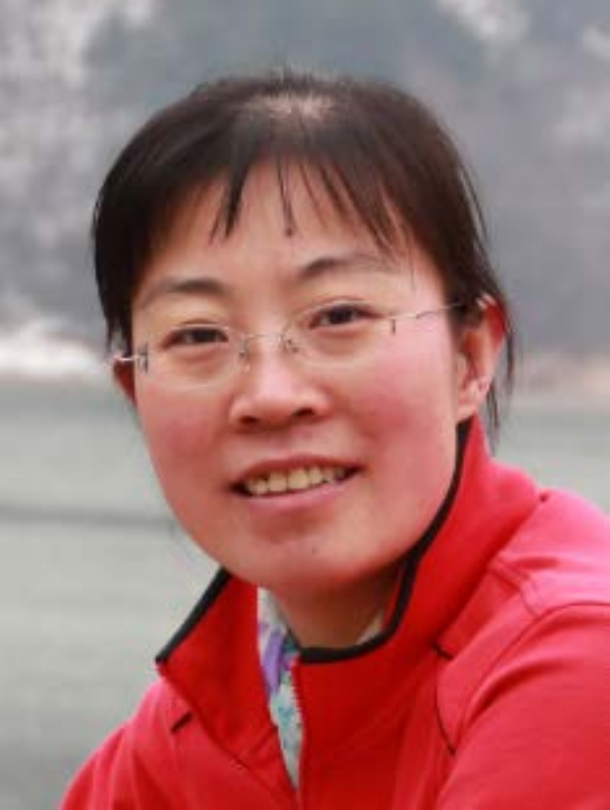}}]{Chunxia~Zhang} received her Ph.D degree in Applied Mathematics from Xi'an Jiaotong University, Xi'an,  China, in 2010. Currently, she is a Professor in School of Mathematics and Statistics at Xi'an Jiaotong University. She has authored and coauthored about 30 journal papers on ensemble learning techniques, nonparametric regression, etc. Her main interests are in the area of ensemble learning, variable selection and deep learning.
\end{IEEEbiography}

\begin{IEEEbiography}[{\includegraphics[width=0.9in,height=1.25in,clip]{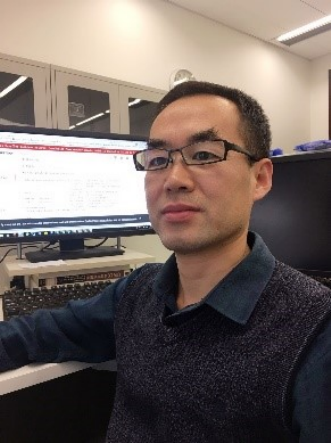}}]{Zhiyu~Yang} received his master degree from the Department of Computer in North China Electric Power University in 2002. Currently, he is a senior engineer in the Geophysical Technology Research Center of Bureau of Geophysical Prospecting (BGP). He is mainly engaged in research and development work of processing system. His current research interests are algorithm optimization and high performance computing.
\end{IEEEbiography}

\begin{IEEEbiography}[{\includegraphics[width=0.9in,height=1.25in,clip]{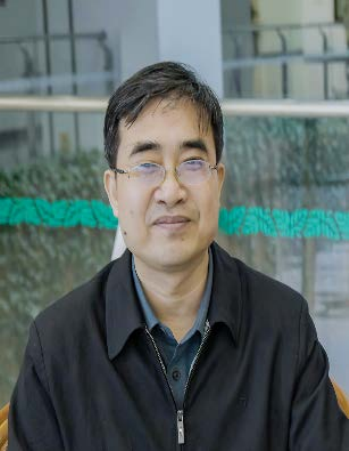}}]{Weifeng~Geng} received his master degree from the College of Geophysics in China University of Petroleum-Beijing in 2002. Currently, he is a senior engineer in the Geophysical Technology Research Center of Bureau of Geophysical Prospecting (BGP). He has been engaged in research of seismic data processing for a long time. His main research interests are VSP (vertical seismic profile) data processing, seismic velocity analysis and high resolution processing.
\end{IEEEbiography}

\end{document}